\documentclass{article}
\usepackage{iclr2027_conference,times}

\usepackage{amsmath,amssymb}
\usepackage{booktabs}
\usepackage{tabularx,array}
\usepackage{xcolor}
\usepackage{graphicx}
\usepackage{hyperref}
\usepackage{xurl}
\hypersetup{hidelinks,pdfauthor={Fei Ding}}
\graphicspath{{figures/}}
\newcolumntype{Y}{>{\raggedright\arraybackslash}X}

\title{Edit-Induced Dependency Mismatch:\\ Propagation Omissions, Recovery, and\\ Evaluation Blind Spots in Coding Agents}
\author{Fei Ding\thanks{Correspondence to Fei Ding,
  \href{mailto:dignfei@gmail.com}{\texttt{dignfei@gmail.com}}.}\\Alibaba Group}
\iclrfinalcopy

\begin{document}
\addtocontents{toc}{\protect\setcounter{tocdepth}{-1}}
\maketitle

\begin{abstract}
During iterative code modification, coding agents may continue to rely in subsequent reasoning on outdated dependency judgments whose text remains fixed in the context, leading to errors in subsequent reasoning and leaving consumer defects undetected by the original task evaluations, thereby exposing blind spots in the tested benchmarks.
We study actual agent execution trajectories collected on benchmark tasks: \texttt{deepseek-v4-flash} runs on DeepSWE and SWE-bench Verified, supplemented by Qwen3.8-Flash cases on FeatureBench.
Focusing on edit-induced consumer obligations, a full-corpus blinded human audit of the DeepSWE runs identifies terminal mismatches, including diagnoses whose task scope remains disputed.
Following supplementary execution checks, all retained DeepSWE terminal events have execution support.
The audit also shows that recovery within a trajectory can coexist with terminal omissions in the same run.
The process analysis links the presentation of relevant consumer code to explicit dependency judgments and subsequent synchronization actions recorded in the trajectories.
An explicit dependency-review prompt reduces terminal omissions under the tested conditions, supporting re-derivation of consumer obligations after edits.
Local replays of selected DeepSWE cases show that synchronizing the diagnosed consumers restores the target behavior while the listed controls continue to pass.
Cross-benchmark execution counterexamples show that the original task evaluations do not fully detect edit-induced dependency mismatches: patches pass those evaluations, yet additional execution checks confirm unintended defects along existing consumer paths.
These findings support measuring task scores, terminal propagation states, and recovery separately.
\end{abstract}

\section{Introduction}
\label{sec:introduction}

Repository-level coding agents complete software engineering tasks through repeated reading, editing, and testing
\citep{NEURIPS2024_5a7c9475,ICLR2025_a4b6ad6b}.
An edit not only introduces new code but can also impose new obligations on unchanged code:
when a constructor gains a field, an existing transformer must preserve it; when a query return value gains a wrapper, other execution paths must recognize it.
Such mismatches can cause the current task evaluation to fail, or escape evaluation and leave unintended defects in patches that receive passing scores.

\begin{figure}[t]
\centering
\includegraphics[width=\linewidth]{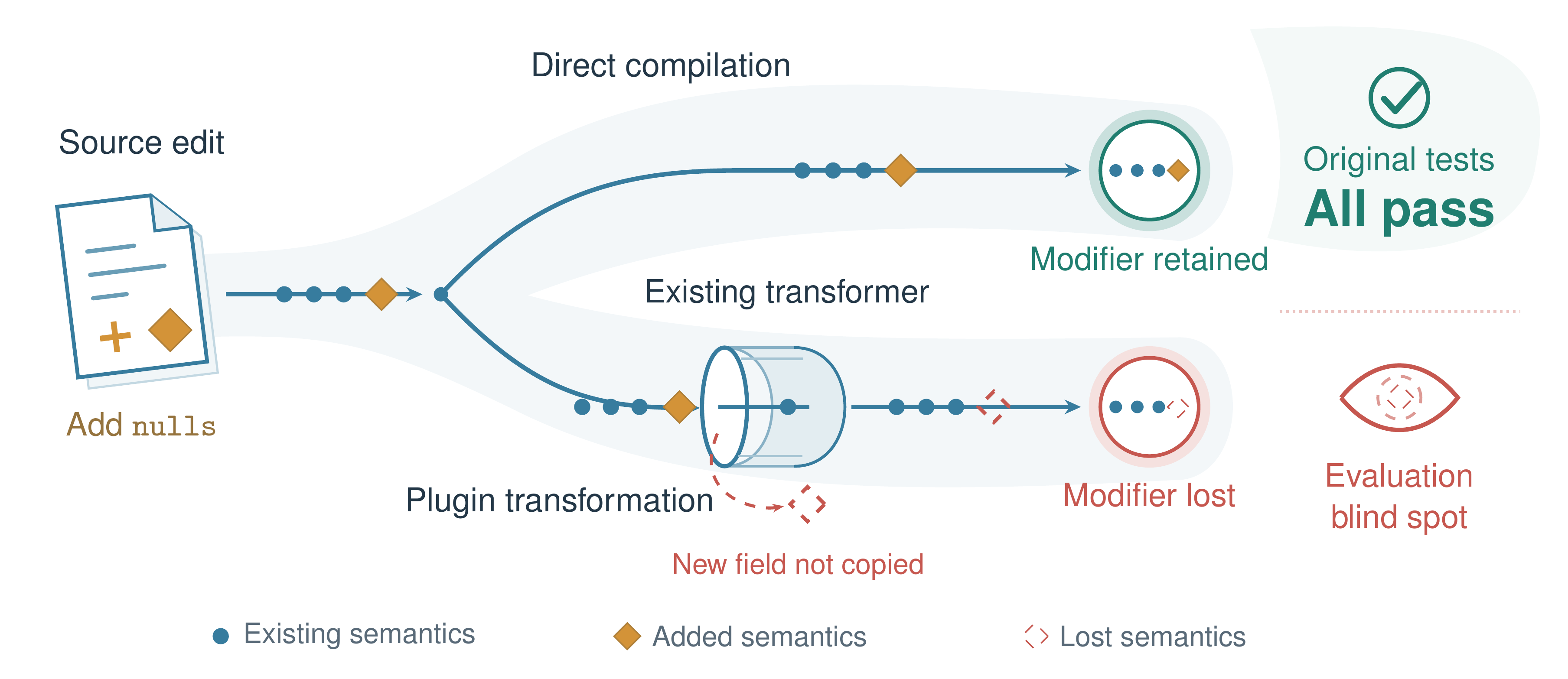}
\caption{Graphical abstract: a passing score does not guarantee complete edit propagation. In Kysely run \texttt{wq82S5C}, the added \texttt{nulls} field survives direct compilation but is not copied by an existing node transformer, losing the modifier on plugin paths despite all original tests passing. Solid diamonds denote added semantics, dashed diamonds their absence, and blue dots existing semantics. The branches illustrate consumer relations; the evaluation marker reports the original run score, not a complete test-coverage map.}
\label{fig:propagation}
\end{figure}

In Query, the infinite-query path fails to consume a new return wrapper, causing the original task evaluation to fail.
Synchronizing the shared consumer entry point makes all original evaluation tests pass while preserving ordinary-query restoration and new-request controls.
In contrast, Kysely passes all original tests, yet its new NULLS modifier is lost in an existing transformer.
Direct compilation preserves the modifier while plugin transformation loses it; adding \mbox{\texttt{nulls: node.nulls}} restores the behavior and preserves the original evaluation (Figure~\ref{fig:propagation}).
These cases expose consumer gaps both detected and missed by the original evaluations.

We call this phenomenon \textbf{edit-induced dependency mismatch}.
Our focus is on consumer obligations introduced by an agent's own edits and existing consumers that fail to meet them.
This phenomenon has two characteristics. First, the mismatch concerns a program relationship: unchanged consumer source code does not guarantee that its behavior remains appropriate for new inputs.
Second, the mismatch has a state over the course of execution: a gap may be repaired later, remain in the final patch, or surface only under additional checks of composed behavior.
Studying only failing runs misses gaps among passing runs; examining only final patches obscures events that have already been repaired.

Change-impact analysis and dependency planning have long studied how to propagate code edits
\citep{BA29120744,Bairi_2024}.
We connect edits in naturally occurring runs, specific consumers, intermediate recovery, and final evaluations into verifiable evidence.
Similar patches may arise from localization failures, code misunderstanding, preservation of old behavior, or omissions of historical information;
program inconsistencies can be tested through execution, whereas cognitive attribution requires different evidence.
We further examine a specific process pattern: a dependency judgment $A$ retained in context is not correctly updated to reflect the code state and consumer obligations, yet continues to guide reasoning during later review.
The trajectory analysis links code presentation, explicit judgments, and subsequent synchronization actions; a dependency-review intervention tests the practical value of re-deriving these obligations (Section~\ref{sec:cases}; Appendix~\ref{app:dependency-review}).

We organize the study around four questions:
\textbf{RQ1}: How can propagation omissions be identified and distinguished from ordinary implementation errors?
\textbf{RQ2}: How do terminal omissions, recovery during the trajectory, and original evaluation outcomes relate to one another?
\textbf{RQ3}: With the model-generated patch held fixed, can synchronizing the diagnosed consumers restore the corresponding behavior?
\textbf{RQ4}: What dependency judgments and validation behaviors are observable after edits, and can explicit dependency review reduce terminal omissions?
The study comprises the following components:
an operational definition based on an ``edit--obligation--consumer--diagnosis'' chain;
a complete run-level audit with cross-tabulations relating terminal omissions, recovery, and original evaluation outcomes, supplemented by terminal and recovery cases from another model on FeatureBench;
targeted DeepSWE and SWE-bench Verified replays that hold patches and inputs fixed, test the behavioral effects of local synchronization, and confirm consumer defects in originally passing patches;
and a process analysis linking code presentation, explicit dependency judgments, and subsequent synchronization actions, together with a dependency-review intervention testing the practical value of re-deriving consumer obligations after edits.

\noindent\textbf{Contributions}
\begin{enumerate}
\item \textbf{A systematic empirical diagnosis and measurement study.} \textbf{The study uses event-level consumer obligations as its unit of analysis, connecting the formation, subsequent reuse, and re-examination of explicit dependency judgments with consumer synchronization outcomes, recovery within trajectories, and task scores. It validates specific program consequences through local synchronization replays and tests the practical effects of reassessing these obligations through targeted dependency-review experiments.}

\item \textbf{New empirical insights}:
\begin{enumerate}
\renewcommand{\labelenumii}{\arabic{enumii}.}
\item \textbf{Process insight:} After a code edit, prior dependency judgments retained in the context may continue to influence subsequent reasoning. In the focal trajectories, the model accepted the same judgment again during later review and drew an incorrect inference. The issue involves \textbf{existing judgments not being effectively updated}.
\item \textbf{Intervention insight:} Under the tested conditions, \textbf{the focus of review matters}. The terminal-omission rate is 9.6\% under general correctness/test checking and 2.4\% with targeted re-derivation of dependency relationships and existing judgments, compared with 10.1\% for the original archived baseline. This provides concrete practical evidence for updating dependency judgments.
\end{enumerate}

\item \textbf{A concrete workflow for improving coding agents.} After each code edit, we explicitly ask the agent to compare the code before and after the edit and check, one by one, which prior judgments have changed as a result. We ask it to re-derive and update those judgments and write them into the current context to reduce the influence of old judgments on subsequent reasoning. Under the same rule for triggering review after each edit, we compare this targeted review with general correctness and test checking. The results support the overall effectiveness of the targeted review.

\item A concrete empirical contribution on evaluation blind spots: \textbf{We provide executable counterexamples in which patches pass all original task evaluations yet retain semantic defects along existing consumer paths, and validate the behavioral consequences of specific consumer relations through local synchronization controls. In Kysely, for example, both the original patch and the synchronized repair pass all 276 original evaluation checks, while the same additional consumer-path checks improve from 0/6 to 6/6, demonstrating a semantic gap that the original score fails to distinguish.}
\end{enumerate}

\suppressfloats[t]

\section{Study Focus and Diagnostic Boundaries}
\label{sec:definition}

\subsection{From Source Edits to Consumer Obligations}

For each event, we record four pieces of evidence, $e=(\Delta,O,C,D)$:
$\Delta$ is a new representation, interface, state, or configuration already written into the source;
$O$ is an adaptation obligation introduced by the edit and supported by task requirements or existing program semantics;
$C$ is a specific consumer that handles the relevant input;
and $D$ is a diagnosis showing that this consumer fails to meet the obligation.
For example, in Kysely, $\Delta$ is the new \texttt{nulls} field,
$O$ is the requirement for node transformations to preserve user-specified modifiers,
$C$ is an existing aggregate-node transformation branch, and $D$ is the difference in attributes and SQL before and after transformation.
A field's optional status in the type system does not eliminate its semantic obligation in a given invocation.

Table~\ref{tab:decision} provides a unified decision table: the edit must have been applied, and the consumer must predate the patch and be distinct from newly written logic and the modified function's own contract;
the obligation must be supported by an explicit task statement or an existing contract that still applies, and reachability must specify an entry point, input/state, branch conditions, and consumer chain.
We exclude a relationship if it clearly fails to meet these conditions; if the obligation or path cannot be established, we retain it as unresolved rather than judging it incorrect merely because it differs from the reference patch.
When the obligation and diagnosis are established but the triggering scenario's inclusion in task scope remains disputed, we retain the diagnosis and record a separate scope flag.
Diagnostic evidence connects execution output to a concrete source-level consumer chain. Supplementary execution support now covers all retained terminal diagnoses; source-level derivations remain part of their diagnostic basis. An existing path in the same file can also qualify when it has a distinct processing role.
Detailed decision rules and boundary cases appear in Appendix~\ref{app:decision}.

\subsection{Terminal States, Recovery, and Program-Level Confirmation}

A \textbf{terminal propagation omission} means that the corresponding gap remains in the final patch;
a \textbf{recovery event} means that a distinct behavioral failure of an existing consumer occurs in the trajectory and is subsequently repaired.
A single run may recover from one event while leaving another unresolved.
Wink monitors coding-agent trajectories and supplies asynchronous guidance to recover from misbehaviors~\citep{nanda2026winkrecoveringmisbehaviorscoding}.
Temporary inconsistencies between planned consecutive edits, migration-time compilation diagnostics, and synchronization of generated files or test snapshots alone fall outside this definition.
Alongside $e=(\Delta,O,C,D)$, we record each identified event's terminal or recovered state. Terminal events also carry evidence $E$ and scope $S\in\{\mathrm{undisputed},\mathrm{disputed}\}$ labels; all retained terminal events now have $E=\mathrm{exec}$ after supplementary execution support (Appendix~\ref{app:aggregation}).

\textbf{Program-level replay confirmation} further fixes the original final patch and check inputs
and compares behavior before and after local synchronization. The local program diagnosis is supported only if the target behavior is restored and the listed controls continue to pass.
Repairs are restricted to the diagnosed consumer set and supporting locations needed to maintain its shared invariants,
while preserving the new functionality, execution paths, check inputs, and assertions.
The repair set may span multiple consumer locations; we assess its local behavioral effects through target recovery on the listed inputs, control checks, and changes in non-target behavior.

\subsection{Relationship to Prior Work}

Surveys and empirical trajectory studies characterize software-engineering agents and their success and failure behaviors~\citep{liu2025largelanguagemodelbasedagents,majgaonkar2025understandingcodeagentbehaviour}.
SWE-bench provides execution-based tasks, while SWE-agent and OpenHands provide agent frameworks
\citep{ICLR2024_edac78c3,NEURIPS2024_5a7c9475,ICLR2025_a4b6ad6b};
RepoCoder, RepoGraph, and Agentless study retrieval, code structure, and repair workflows, respectively
\citep{zhang-etal-2023-repocoder,ICLR2025_4a4a3c19,Xia_2025}.
DeepSWE examines incomplete patches accepted by original tests~\citep{huang2026deepswemeasuringfrontiercoding}; SWE Refactor Bench separately evaluates migration completeness and behavioral correctness~\citep{hong2026swerefactorbenchcoding}.

CodePlan coordinates repository edits through dependency analysis and adaptive planning~\citep{Bairi_2024}.
RefactorBench supports stateful reasoning with updated summaries of accumulated edits~\citep{gautam2025refactorbench}, while STALE studies implicit invalidation of stored beliefs about user state by new observations~\citep{chao2026stale}.
PlanFence separates shared-input freshness from plan validity in distributed agent workflows, checking declared dependencies before actions~\citep{chen2026planfence}.
Our audit links agent-authored edits, existing consumers' semantic obligations, terminal omissions, recovery, and original scores in actual coding trajectories.
Selected local replays test program consequences; focal cases trace explicit dependency judgments persisting through later review, and the intervention tests re-deriving consumer obligations.
Detailed comparisons with refactoring and feature-development benchmarks, together with additional repository-level approaches, appear in Appendix~\ref{app:additional-related-work}.

\section{Data and Audit Method}
\label{sec:data}

\begin{figure}[t]
\centering
\includegraphics[width=\linewidth]{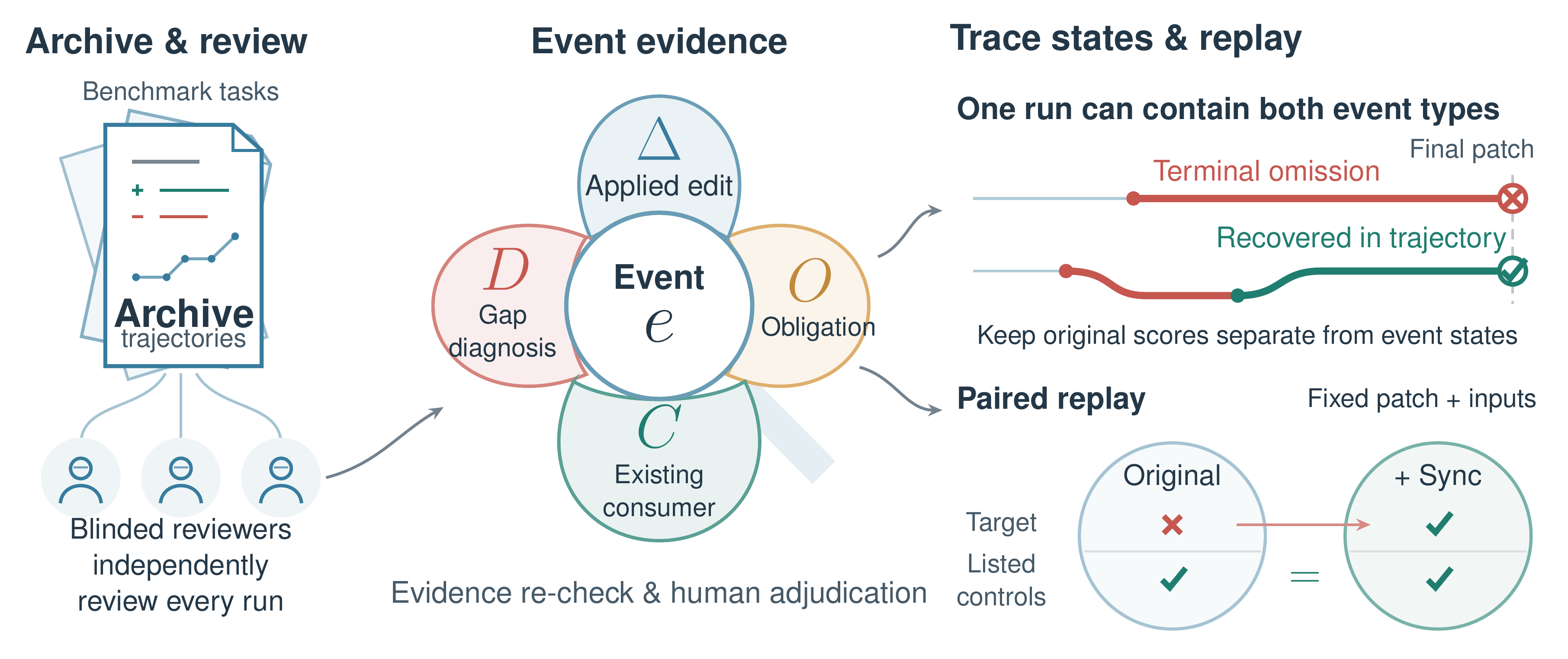}
\caption{Retrospective human audit and local validation. Reviewers independently audit every run in the DeepSWE main corpus, establishing $e=(\Delta,O,C,D)$: applied edit, adaptation obligation, existing consumer, and gap diagnosis. Evidence re-examination and discussion adjudicate events; original score summaries and prior labels are blinded. Terminal omissions and recovery are recorded per event and may coexist within one run. Terminal events additionally retain execution support and scope-dispute labels. Original evaluation outcomes remain separate. Paired replays fix the original patch and inputs, comparing behavior before and after local consumer synchronization: targets change from fail to pass while listed controls pass in both versions. Confirmation is limited to the tested relations and inputs.}
\label{fig:audit-method}
\end{figure}

Figure~\ref{fig:audit-method} summarizes the audit from archived evidence to event identification, state recording, and local program validation.

\subsection{Fixed Run Corpus}

The tasks come from the DeepSWE benchmark and its public repository
\citep{huang2026deepswemeasuringfrontiercoding,datacurve2026deepswe}.
We define the complete DeepSWE corpus using the unique run identifiers in the archived score table.
The main corpus contains $2{,}879$ runs across $64$ tasks; these counts describe the full DeepSWE audit and exclude supplementary benchmarks.
Only officially scored runs enter the corpus; planned repetitions and additional archived directories do not expand the denominator.
Every included record links to a final patch, a complete normalized execution trajectory, and the original evaluation outcome.
Table~\ref{tab:data} reports the corpus and run settings.

\begin{table}[t]
\caption{Main and supplementary corpora and run settings. The full DeepSWE audit and cross-benchmark analyses are reported separately and do not share a pooled denominator.}
\label{tab:data}
\centering\small
\begin{tabularx}{\linewidth}{>{\raggedright\arraybackslash}p{.24\linewidth}Y}
\toprule
Item & Archived facts and measurement units \\
\midrule
DeepSWE main corpus & $2{,}879$ runs, $64$ tasks, $55$ repository snapshots \\
Scored-set boundary & $2{,}880$ runs planned; $1$ did not enter the officially scored set \\
Original evaluation outcomes & $1{,}286$ passing; $1{,}593$ failing; original values retained \\
Final patches & Median of $9$ modified paths; $2{,}868$ runs modify at least two files; $11$ patches are empty \\
Coding-run stack & \texttt{deepseek-v4-flash}; \texttt{mini-swe-agent 2.4.6}; Harbor \texttt{0.21} \\
Original conditions & \texttt{bare}, \texttt{inject-file}, \texttt{guided}; $15$ runs planned per task per condition \\
Audit procedure & Three independent full-corpus blinded human reviews plus a separate independently sampled blinded audit; final labels determined by humans \\
Pre-discussion unanimous agreement & Primary category $97\%$; complete event tuples $89\%$, covering terminal and recovery events \\
SWE-bench Verified extension & $500$ final trajectories across $12$ repositories; $462$ originally passing and $38$ failing; \texttt{deepseek-v4-flash}, SWE-agent~1.1.0 \\
FeatureBench extension & Qwen3.8-Flash; confirmed-case analysis restricted to submitted answers; counts and outcomes in Table~\ref{tab:featurebench-full100} \\
Statistical units & Runs, task coverage, diagnosed events, and within-task consumer relation types, reported separately \\
\bottomrule
\end{tabularx}
\end{table}

The full model identifier is \texttt{openai/deepseek-v4-flash}, whose prefix is a provider tag used by the execution framework.
The original conditions provide no repository index, provide an index, and add a usage protocol to the index, respectively; these are run-level settings.
Modified-path counts measure patch breadth across source, tests, documentation, and configuration; consumer relations are counted separately.
Parameters follow the default benchmark configuration.

\subsection{Full-Corpus Blinded Review and an Independently Sampled Blinded Audit}

We used annotation guidelines preregistered before exploratory analysis. Three reviewers with experience in large language models and code analysis each independently reviewed the complete DeepSWE corpus under blinded conditions, establishing edit--obligation--consumer--diagnosis evidence chains from task requirements, final patches, execution records, and trajectories.
The full-corpus review hid original score summaries, existing audit labels, and the other reviewers' rationales while retaining execution evidence needed for diagnosis.
Before discussion, unanimous agreement was measured separately for the primary category and complete event tuples $e=(\Delta,O,C,D)$, covering both terminal and recovery events (Table~\ref{tab:data}).
Disagreements were resolved through evidence re-examination and discussion.
Primary labels, terminal states, and events were recorded separately; ``unresolved'' is the reviewers' agreed classification for insufficient diagnostic evidence.

In addition to the full-corpus blinded review, we conducted an independently sampled blinded audit with the original task-evaluation score summaries and existing audit labels hidden.
Reviewers judged cases independently using task requirements, source code, original model patches, and relevant execution evidence; initial annotations were fixed before unblinded comparison and adjudication of disagreements.

Aggregate statistics describe the original full-corpus human audit. Two independent external experts ultimately confirmed $466$ of the $473$ originally diagnosed runs (Appendix~\ref{app:external-review}).
Local replays test the behavioral effects of synchronizing selected consumer relations.
Appendix~\ref{app:protocol} describes the audit chronology and final criteria; Appendix~\ref{app:artifact} identifies the submitted data and diagnosis records.

\subsection{Program State and Model-Visible Information}

Program-level attribution and process interpretation require different evidence.
Source code, patches, and raw execution outputs can support judgments about program behavior;
claiming that the model ``read a consumer,'' however, requires locating the relevant code in the tool messages actually presented at the time.
Complete archived outputs may retain sections omitted from those messages and are used for program analysis.
The focal cases therefore separately record three things: whether relevant information was presented, whether an explicit judgment was made, and whether synchronization and validation were performed.
Explicit judgments provide process evidence at the message level.
Audit sources and final criteria appear in Appendix~\ref{app:protocol}; the submitted-materials index is in Appendix~\ref{app:artifact}.

\section{Relationships among Propagation Omissions, Recovery, and Evaluation Outcomes}
\label{sec:results}

\subsection{Terminal Labels Cover Both Passing and Failing Runs}

For run $r$, let $T_r$ contain its diagnosed terminal events and $A_r=\mathbf{1}[T_r\ne\varnothing]$. For the main corpus of $R$ runs, the run share is $p_{\mathrm{run}}=R^{-1}\sum_{r=1}^{R}A_r$.
Terminal diagnoses occur in both originally passing and failing runs and span distinct tasks and within-task consumer relations (Table~\ref{tab:audit}).
We report the full diagnosis set, including disputed scope, separately from the subset whose terminal events are all undisputed.
These sets account for $16.4\%$ and $10.1\%$ of the same main corpus, respectively; the shares describe this fixed corpus rather than estimating a broader population mismatch rate.
After supplementary checks, every retained terminal event has execution support; an execution-support filter therefore no longer reduces either set.
Appendix~\ref{app:aggregation} gives the aggregation rules and full evidence distribution; Appendix~\ref{app:artifact} links diagnoses to their sources.

\begin{table}[t]
\centering\small
\caption{Terminal diagnoses (including disputed scope), evidence, and recovery after supplementary execution support. All diagnosed runs have execution support. The undisputed subset requires all terminal events in a run to have undisputed scope. Original evaluation outcomes are retained.}
\label{tab:audit}
\begin{tabularx}{\linewidth}{Yrrrrrr}
\toprule
\multicolumn{7}{l}{\textbf{(a) Terminal diagnoses by scope, evidence, and original outcome}} \\
\shortstack[l]{Original\\outcome} & \shortstack[r]{All\\runs} & Diagnosed & \shortstack[r]{Any\\exec.} & \shortstack[r]{Static\\only} & \shortstack[r]{All events\\undisputed} & \shortstack[r]{Undisputed\\+ execution} \\
\midrule
Passed & $1{,}286$ & $146$ & $146$ & $0$ & $75$ & $75$ \\
Failed & $1{,}593$ & $327$ & $327$ & $0$ & $215$ & $215$ \\
\midrule
Total & $2{,}879$ & $473$ & $473$ & $0$ & $290$ & $290$ \\
\midrule
\multicolumn{7}{l}{Terminal diagnoses: 37 tasks / 85 consumer relation types / 473 runs / 546 events} \\
\multicolumn{7}{l}{Run shares: all diagnoses $16.4\%$; all events undisputed $10.1\%$; same main-corpus denominator} \\
\multicolumn{7}{l}{Task coverage of terminal diagnoses: $57.8\%$} \\
\bottomrule
\end{tabularx}
\smallskip
\begin{tabularx}{\linewidth}{Yrrr}
\toprule
\multicolumn{4}{l}{\textbf{(b) Terminal status and recovery events}} \\
Terminal status & \shortstack[r]{With recovery\\events} & \shortstack[r]{No recovery\\event recorded} & Total \\
\midrule
Terminal diagnosis & $83$ & $390$ & $473$ \\
Other (including unresolved) & $287$ & $2{,}119$ & $2{,}406$ \\
\midrule
Total & $370$ & $2{,}509$ & $2{,}879$ \\
\midrule
\multicolumn{4}{l}{Recovery event coverage: 36 tasks / 415 events} \\
\bottomrule
\end{tabularx}
\end{table}

Execution support includes original tests, in-trajectory probes, local repair replays, and supplementary checks.
Local repair replays remain a selected repair-confirmation subset; propagation omissions in the failed pool may also coexist with other errors.

\paragraph{Task concentration and scope sensitivity.}
Terminal annotations are concentrated in a subset of tasks (Table~\ref{tab:relations}).
Relations are deduplicated by task, producer-side semantic change, adaptation obligation, and existing consumer path; a single run may involve multiple relations.
Repeated runs describe recurrence of these relations; neither run counts nor relation counts represent independent samples.
The manual scope review distinguishes runs with only undisputed terminal events, runs with both disputed and undisputed events, and runs with only disputed events.
Disputed and undisputed events in mixed runs are recorded separately; only runs whose terminal events are all undisputed enter the strict scope subset.
Appendix~\ref{app:distribution}~provides the scope groups, evidence types, and counting conventions.

\subsection{Recovery and Terminal Omissions Can Coexist}

Runs with recovery events can still retain other terminal omissions (Table~\ref{tab:audit}(b)).
Recovery and terminal omissions are counted from their respective event records; absence of either annotation does not establish absence of defects.
Together, these records characterize the intermediate and final states of change propagation,
while natural recovery trajectories provide additional process evidence on when gaps were discovered and how synchronization was performed.

Among runs failing the original evaluation, specification/interface errors and local algorithmic errors are the most common primary labels. Appendix~\ref{app:taxonomy} reports the complete classification, and Appendix~\ref{app:decision} details the diagnostic boundaries.

\section{Propagation Paths and Discovery Processes}
\label{sec:cases}

\subsection{Terminal Omission and Delayed Recovery of the Same Type of Gap}

In Kysely, \texttt{wq82S5C} (\texttt{inject-file}) incorporated \texttt{nulls} into the node definition, construction interface, and SQL compilation,
but omitted the aggregate-node branch of the existing \texttt{OperationNodeTransformer}.
This branch still copied the old fields one by one, allowing it to produce a type-valid new node that lost user-specified semantics.
This terminal case records a program-level propagation gap under a condition in which a repository index was provided.

In an independent \texttt{inject-file} run on the same task, \texttt{5WQPAYr}, the original agent discovered and repaired the same transformer omission after initially reporting that its own tests passed (Appendix~\ref{app:kysely-recovery}).

\subsection{A Shared Entry Point Does Not Guarantee a Shared Consumer Path}

In Query, \texttt{MYCfNHF} (\texttt{bare}) introduced a return wrapper carrying recovery state
and unwrapped it in the internal \texttt{fetchFn} for ordinary queries.
Infinite queries, however, replaced \texttt{context.fetchFn} through a behavior extension, bypassing this logic.
The model inferred identical recovery behavior because both used the outer \texttt{Query.fetch},
but a shared outer entry point does not guarantee that internal results pass through the same consumer.
The original record reports passing existing tests but a failing new test targeting the recovery state of infinite queries (Table~\ref{tab:original-regression}).
This case grounds the abstract notion of a ``dependency omission'' in a consumer operation bypassed by a replaced branch.

In termenv, \texttt{5FGCqii} (\texttt{bare}) passed \texttt{preserveResets} from the output object to style objects and templates,
but ordinary rendering retained the old concatenation logic, leaving the new state consumed only on the truncation path.
The model actively requested \texttt{cat style.go}, and the displayed response contained the old \texttt{Styled} in full.
After the edit, the archived reasoning text states, ``The preserveResets field only affects Truncate, not Styled.''
A later reasoning record accepts unchanged template-style output on the premise that ``Only Truncate uses preserveResets. That's fine.''
Here, the explicit judgment is formed after the new state is introduced and persists through later review, while the ordinary-rendering obligation remains unmet.
Code presentation and subsequent review therefore do not ensure that the judgment is updated to reflect the current program state.
Appendix~\ref{app:prior-judgment} locates the code presentation and judgment records and distinguishes their source fields.
Once the consumer began responding to the state, the truncator's existing logic for reopening styles also required coordination.
These two cases show partial integration of a return representation and a rendering state, respectively,
and demonstrate that synchronization obligations must be checked along actual execution paths rather than inferred from a shared object or function name.

The supplementary eicrud case links a tentative pre-edit decision, a new server constraint, and retention of that decision after the edit; its task scope is disputed and the consumer conflict is statically derived, so it serves as supplementary process evidence (Appendix~\ref{app:prior-judgment}).

\subsection{Explicit Awareness of a Gap May Still Leave Synchronization Incomplete}

In Helm, \texttt{BRvPrLE} (\texttt{bare}) added merge strategies for arrays,
but passed only the command-line strategy when reusing configuration during an upgrade, omitting Chart annotations.
The model explicitly discussed this gap, then concluded that the subsequent merge stage would compensate for it.
In fact, the subsequent stage could not recover array elements already discarded by the preceding stage.
This record contains an explicit discussion of the gap, an incorrect judgment about a subsequent stage, and a final program state that remained unsynchronized.

\paragraph{Re-examining retained dependency judgments.}
Both review groups resume from matching checkpoints with the same starting context, code state, and token-budget cap. Both receive a review prompt after each edit. Archived normal-continuation runs serve as the baseline, and all groups use the same undisputed-scope criterion.
Explicit dependency review leaves fewer terminal omissions than either comparison condition (Table~\ref{tab:dependency-review}).
The corresponding run shares are $10.1\%$ for normal continuation, $9.6\%$ for general checking, and $2.4\%$ for explicit dependency review.
This supports re-deriving dependency obligations after edits as a practical intervention; its setup and limits, descriptive process statistics, and a supplementary before/after judgment sequence appear in Appendices~\ref{app:dependency-review}, \ref{app:process-review}, and \ref{app:prior-judgment}.

\section{Execution Checks of Consumer Relations and Cross-Benchmark Counterexamples}
\label{sec:replay}

\subsection{Fixed Patches, Fixed Checks, and Local Synchronization}

We selected five locally executable DeepSWE cases covering node fields, return wrappers, rendering state, configuration sources, and error-value representations (Table~\ref{tab:replays}).
For each, we fixed the repository starting point, original model patch, and check program, then compared the original patch with local consumer synchronization.
Repairs followed task behavior and consumer paths without adopting the reference patch.

Table~\ref{tab:replays}~reports the actual execution results.
Each row corresponds to one run, with multiple assertions covering different paths; parent and child pass events in the original suite are reported according to the test records.
Program-level confirmation is based on restoration of the target behavior under fixed inputs and preservation of the listed controls, linked to specific terminal events.
The complete repair process and execution scope are provided in Appendix~\ref{app:replaydetails}.

\begin{table}[t]
\caption{Local program replays of five fixed patches. The target column reports the number of passing checks in the same set before and after repair; the preserved column lists checks that pass in both versions. Each row represents one trial; individual checks are not treated as independent samples.}
\label{tab:replays}
\centering\small
\begin{tabularx}{\linewidth}{p{.17\linewidth}Yp{.17\linewidth}>{\raggedright\arraybackslash}p{.25\linewidth}}
\toprule
Task/trial & Local synchronization & Target passes & Preserved passing checks \\
\midrule
Kysely\newline\texttt{wq82S5C} & Transformer copies the \texttt{nulls} field (1 file, 1 added line) & $0/6\to6/6$ & $2/2$ controls; original evaluation $276/276$; source and test type checks \\
Query\newline\texttt{MYCfNHF} & Shared result entry point unwraps the restoration marker (1 file, net 4 added lines) & $0/4\to4/4$ & $8/8$ ordinary restoration and fresh-request controls \\
termenv\newline\texttt{5FGCqii} & Renderer reads preservation state; truncation avoids reopening styles twice (2 files) & $0/7\to7/7$ & $12/12$ controls; original repository tests \\
Helm\newline\texttt{BRvPrLE} & Configuration merge reads annotation policies and retains CLI precedence (2 files) & $0/5\to5/5$ & $48/48$ controls and regression leaf tests \\
Expr\newline\texttt{7uAWePB} & \texttt{string()} preserves error values and reads their messages (2 files) & $0/6\to6/6$ & $9/9$ controls; original repository tests \\
\bottomrule
\end{tabularx}
\par\smallskip\raggedright\footnotesize
Kysely belongs to \texttt{inject-file}, and the other four trials belong to \texttt{bare}; these groups indicate the index-provision condition, while visibility of specific code is checked against each trajectory.
Helm's 48 checks comprise 9 local controls, 5 existing action regression leaf tests, and 34 util leaf tests.
Original evaluation outcomes and local replays are recorded separately; Appendix~\ref{app:replaydetails} details the full execution scope and changes in non-target behavior.
The final Query repair includes explicit generic type annotations; Table~\ref{tab:original-regression} reports the original evaluation reruns.
\end{table}

\subsection{Repairable Semantic Loss Despite Passing All Tests}

The Kysely replay applied the original patch to the fixed starting point \texttt{91cf3733},
constructed queries through the public API, and inspected AST attributes and SQL compilation output.
Both \texttt{ignoreNulls()} and \texttt{respectNulls()} retained their semantics on the direct path;
after passing through the new \texttt{SimplifyFramePlugin}, the generic node transformer, or the existing
\texttt{WithSchemaPlugin}, both modifiers disappeared.
For example, \texttt{first\_value(...)\ ignore nulls\ over(...)}
became \texttt{first\_value(...)\ over(...)}.
Copying the field restored the transformation checks, while the direct-path controls continued to pass.

The benchmark validator applied \texttt{test.patch}; the original evaluation comprised new acceptance tests (F2P) and existing regression tests (P2P), all of which passed (Table~\ref{tab:original-regression}).
These passing results, together with the additional checks, localize a blind spot on composed paths; the original evaluation is counted separately from the agent's own tests and our probes.
In supplementary reruns, both the original patch and the consumer repair passed the same original evaluation tests,
as well as type checks for the production source in ESM/CJS and the Node test code (Appendix~\ref{app:replaydetails}).
These results confirm a semantic gap in composition within this successful patch that can be repaired by synchronizing the omitted field.

Local synchronization restores the targeted behavior and preserves the listed controls. The Expr repair also changes the representation of environment-supplied error values. Appendix~\ref{app:replaydetails} reports these checks and changes in detail.

\subsection{Failure Diagnosis and Successful-Patch Counterexamples on SWE-bench Verified}
\label{sec:sweverified}

We also analyzed SWE-bench Verified runs using the same \texttt{deepseek-v4-flash} model. Appendix~\ref{app:sweverified} describes the archived run settings and diagnoses of originally failing patches.

We purposively selected cases from the successful pool for execution replays of the original patches: the old \texttt{:any:} path failed to recognize new entry keys in Sphinx~7440;
the old product reduction discarded the delta range in SymPy~12419; and variable attributes incorrectly entered the coordinate-merging path in xarray~6461.
Local synchronization restored the targets in these cases, while the executed controls related to the original tasks continued to pass (Table~\ref{tab:sweverified-replays});
the execution and regression scope are provided in Appendix~\ref{app:sweverified}.

On FeatureBench, submitted Qwen3.8-Flash runs provide execution-supported examples of terminal omissions and within-run recovery (Appendix~\ref{app:featurebench}). Appendix~\ref{app:featurebench-modes} covers first attempts, including timeouts, and Appendix~\ref{app:codex-featurebench} reports the separate Codex audit.

\section{Discussion and Limitations}
\label{sec:discussion}

\paragraph{Task evaluation and change impact.}
Execution counterexamples from DeepSWE and SWE-bench Verified reveal consumer defects undetected by the original task evaluations.
This motivates checking consumers' new obligations and preserving evaluation outcomes, terminal states, and recovery records separately.

\paragraph{Program-level and cognitive attribution.}
Local repair checks examine the effects of code on behavior. Our analysis of natural recovery and the original conditions is descriptive.
The dependency-review intervention (Appendix~\ref{app:dependency-review}) tests a specific workflow prompt.
It does not separately identify the cognitive effect of trusting a prior judgment, or the contributions of model reasoning capacity and workflow constraints.
Causal effects of repository indexes, code graphs, and retrieval strategies remain outside this study.

\paragraph{Measurement boundaries and generalization.}
Benchmark studies examine potential model-memory confounds and compare agent harnesses under controlled evaluation protocols~\citep{prathifkumar2025doesswebenchverifiedtestagent,zheng2026clawswebenchbenchmarkevaluatingopenclawstyle}.
We conduct a descriptive audit of the fixed corpus, reporting evidence strata, scope groups, run counts, and task coverage separately, with exploratory groups in Appendix~\ref{app:distribution}.
DeepSWE and SWE-bench Verified use the same model while tasks, budgets, and frameworks vary together.
FeatureBench adds Qwen cases using the same framework as DeepSWE and a separate Codex archive audit; the differing samples and run conditions do not support model rankings or mismatch-rate comparisons.
We selected replay cases for localizability and executability; conclusions concern the tested inputs, consumer repairs, and listed regression checks.
Overall mismatch rates, average repair benefits, and causal effects across frameworks are outside the scope of our estimates.

\section{Conclusion}
\label{sec:conclusion}

Edit-induced dependency mismatch can cause evaluation failures and can also leave consumer defects in patches that pass the evaluation.
The audit identified terminal mismatches supported by execution evidence and recorded task-scope disputes separately.
The process cases further show that code presentation and subsequent review do not ensure updated dependency judgments; the intervention supports explicitly re-deriving consumer obligations after edits.
Targeted replays across benchmarks examined the behavioral effects of local synchronization and exposed propagation defects missed by the original task evaluations,
while Qwen cases on FeatureBench add cross-model evidence of terminal mismatches and recovery.
These findings support separate measurement of task evaluation outcomes, terminal states, and recovery, without estimating broader mismatch rates.

\clearpage
\subsection*{AI Usage Statement}

No AI was used in this work.

\subsection*{Ethics Statement}

Not applicable.

\subsection*{Reproducibility Statement}

The supplementary materials provide all DeepSWE model responses and scores, together with final patches, audit labels, and terminal diagnostic opinions, enabling reproduction of the results reported in this paper.

\clearpage
\appendix
\renewcommand{\floatpagefraction}{.85}
\addtocontents{toc}{\protect\setcounter{tocdepth}{2}}

\begingroup
\renewcommand{\contentsname}{Appendix Contents}
\hypersetup{linktoc=all}
\small
\phantomsection
\pdfbookmark[0]{Appendix Contents}{appendix.contents}
\tableofcontents
\endgroup

\section{Audit Criteria and Version History}
\label{app:protocol}

\subsection{Data Freeze and Provenance}

The earlier audit used the \texttt{deepswe-retrospective-census-v1.0} protocol.
Earlier exploration examined evaluation outcomes in the original failed pool and inspected failure signatures
across $24$ tasks and $896$ failed records; these counts describe the exploration workload.
The archival source for the final statistics is \path{full_audit/canonical/primary.jsonl}; its study-archive manifest confirms complete coverage of all $2{,}879$ runs and successful structural validation.
The submission includes a minimal cohort projection at \path{deepswe_core_evidence_slim/data/final_labels.jsonl}, together with the original scores and patches.
Current terminal-event evidence and scope labels are supplied separately in \path{deepswe_473_terminal_evidence/labels.jsonl}.
Coverage of event-level diagnostic fields is reported separately in Appendix~\ref{app:distribution}.
The archival source file's SHA-256 hash is
\begin{quote}\small\ttfamily
5479ebf42555db8b59e8f441285f8b937\\
becdcbd5ea4d315f7ab9cbfaa060877
\end{quote}
The hash and structural validation verify data provenance and version consistency.

The human annotation guidelines were preregistered before exploratory analysis.
After exploration and refinement of the operational criteria, three human reviewers independently re-reviewed all runs under common criteria to determine the classifications and evidence assignments used in this paper.
A separate independently sampled blinded audit provided additional validation alongside the full-corpus blinded review.
We determine diagnoses through the full human review; the submitted core directory retains run identifiers and source hashes for linking the distributed records.
In the study archive, each audit record links the task, patch, relevant trajectory windows, and available diagnostics,
and preserves alternative explanations and limitations of the materials. Aggregate statistics deduplicate runs by their unique identifiers.

\subsection{Unified Classification Boundaries}

\label{app:decision}
Table~\ref{tab:decision} organizes the retrospective diagnostic criteria applied to the frozen audit into a sequence of checks against the source materials.
Each decision concerns a specific propagation relation; a run's primary category incorporates coexisting diagnoses according to the protocol's priority order.

\begin{table}[htbp]
\caption{Unified decision table for propagation relations. Scope status and evidence level are recorded separately; confirmed exclusions and insufficient materials are treated separately.}
\label{tab:decision}
\centering\small
\begin{tabularx}{\linewidth}{>{\raggedright\arraybackslash}p{.15\linewidth}Y>{\raggedright\arraybackslash}p{.30\linewidth}}
\toprule
Criterion & Required evidence & Treatment when false or indeterminate \\
\midrule
Implemented edit & Baseline, patch, and trajectory from the same run, identifying the actual change $\Delta$ & Plans alone are excluded; missing materials remain recorded as missing \\
Established obligation & Exact task wording or a still-applicable existing contract explaining why the consumer must adapt & Differences from the reference implementation are insufficient; unresolved conflicts in obligations remain unresolved \\
Existing and reachable & An independent consumer present before the patch; a specific entry point, input/state, branch conditions, and consumption chain & Errors within a newly written/rewritten function are classified separately; confirmed unreachability leads to exclusion, while gaps in the path remain unresolved \\
Substantiated diagnosis & An actual erroneous output for the same relation, with a concrete source-level account; source-only diagnoses from the original audit receive supplementary execution support & Evaluation outcomes, test names, and self-reports are insufficient substitutes; cases without a verifiable diagnosis remain unresolved \\
Terminal state/recovery & A gap remaining in the final patch; or a runtime failure of an existing consumer that is subsequently repaired & Repaired gaps are not counted as terminal; migration transients and generated-file synchronization do not automatically count as recovery \\
Scope and aggregation & Scope disputes recorded separately after the obligation and diagnosis are established, followed by separate counts of events and runs & Scope status follows explicit human adjudication; historical missing fields and supplementary decisions are recorded separately \\
\bottomrule
\end{tabularx}
\end{table}

\paragraph{Conflicting obligations and source-level diagnostic checks.}
We identify the specific behaviors supported by the task wording and the existing contract separately, and check their respective scopes of applicability.
The reference patch serves only as an interpretive aid; when conflicting interpretations prevent the expected behavior from being determined, the obligation remains unresolved.
If the obligation and a reachable deviation have been established, but whether the task description covers the input combination remains disputed, we retain the diagnosis and record a scope dispute.
The source-level derivation accompanying execution evidence must specify the entry point, input/state, branch conditions, call/data flow, and erroneous outcome,
and check for guards that block the path or branches that are not connected; functions with the same name, changes across multiple files, or failures in other runs are insufficient substitutes.

\paragraph{Boundary examples.}
The field loss in Kysely \texttt{wq82S5C} meets the inclusion criteria and is supported by a repair replay;
the combination of configuration reinjection and append in Helm \texttt{27jewNv} has supplementary execution support, while whether the rendering combination falls within the task scope remains a separate dispute.
In Go-genai \texttt{JupuHoi}, the retroactive modification of response snapshots occurs within the new accumulator itself; despite an actual failure, it is classified as a local algorithm error rather than propagation.
In Eicrud \texttt{5Xa3tcN}, the existing client's subsequent retrieval combines offset with the new cursor restriction, creating a conflict,
but it could not be established whether the server-side \texttt{\$find} task imposes this obligation for the client's new usage, so propagation remains unresolved.
The latter two cases are excluded from the $473$ runs; their primary labels are retained in the submitted cohort projection.

Primary labels organize records according to the protocol's priority order, while secondary labels and events preserve coexisting diagnoses.
Program evidence and process evidence are used separately: execution-evidence levels follow the recorded behavior, and model-visible code is identified in the messages presented to the model.
Complete archived outputs support program analysis; unresolved relations with insufficient evidence and relations excluded on sufficient evidence are recorded separately.
Our process analysis focuses on the focal cases, linking code presentation, explicit judgments, and synchronization actions to their trajectory locations.

\subsection{Human Review Procedure}

The annotation guidelines were preregistered before exploratory analysis. Full-corpus blinded human review followed refinement of the operational criteria. Three reviewers
each independently reviewed all $2{,}879$ runs under the unified criteria and blinded conditions, covering runs that passed and failed the original evaluation.
For each run, reviewers checked the task requirements, final patch, execution records, and relevant trajectories, and recorded the classification, consumer obligation, diagnostic basis, and alternative explanations.
The full-corpus blinded review hid the original task-evaluation score summaries, existing audit labels, and the other reviewers' labels and rationales while retaining the program evidence needed for diagnosis.
Before discussion in this full-corpus blinded stage, unanimous agreement among all three reviewers on the primary category was $97\%$, with all $2{,}879$ runs as the denominator;
unanimous agreement on the complete event tuple $e=(\Delta,O,C,D)$ before discussion was $89\%$, with event comparisons covering both terminal and recovery events.
The former measures agreement on the run's primary category; the latter requires agreement on all four components of the complete tuple.
Evidence rechecking and discussion resolved all reviewer disagreements; the remaining ``unresolved'' labels denote insufficient diagnostic evidence.
We report both pre-adjudication unanimous agreement rates at the aggregate level as integer percentages.

\paragraph{Independently sampled blinded audit.}
We conducted a separate independently sampled validation in addition to the blinded review of all $2{,}879$ runs.
This stage hid the original task-evaluation score summaries and existing audit labels while retaining task requirements, source code, original model patches, and relevant execution evidence.
Reviewers first made independent judgments and fixed their initial annotations, followed by unblinded comparison and adjudication of disagreements.
The $97\%$ and $89\%$ figures above describe agreement in the full-corpus blinded review; the independently sampled audit is a separate procedure.
This subsection describes the procedure; the submitted files contain the final cohort labels and terminal diagnosis opinions.

\subsection{Independent External Expert Review}
\label{app:external-review}

Two external experts, each with at least five years of experience in software development and code review, independently reviewed the original set of terminal-diagnosis runs.
They had no direct collaborative relationship with the authors and had not participated in the original annotation. The original audit labels and task-evaluation scores were hidden; reviewers inspected the task requirements, patches, consumer paths, and execution evidence.

Table~\ref{tab:external-review} summarizes the review stages. Runs not jointly accepted initially underwent follow-up with additional evidence or corrections.
The final externally confirmed set comprises the runs accepted initially and those confirmed after follow-up; excluded and unresolved runs are outside that confirmed set.

\begin{table}[htbp]
\centering\small
\caption{External expert review of the original terminal-diagnosis set. Counts are runs; rows report successive review stages and the final summary, rather than mutually exclusive categories to be added together.}
\label{tab:external-review}
\begin{tabularx}{\linewidth}{Yr}
\toprule
Stage / outcome & Runs \\
\midrule
Original terminal-diagnosis set & $473$ \\
Initially accepted by both experts & $451$ \\
Requiring follow-up after initial review & $22$ \\
\midrule
Confirmed after follow-up & $15$ \\
Excluded after follow-up & $4$ \\
Unresolved after follow-up & $3$ \\
\midrule
Final externally confirmed set & $466$ \\
\bottomrule
\end{tabularx}
\end{table}

These are run-level external-review outcomes. Review disagreements are distinct from task-scope disputes; an unresolved external review is also distinct from evidence-source or terminal-status uncertainty in the original audit.
The main audit totals and cross-tabulations describe the original audit cohort, while this subsection separately reports its external validation. Revising event totals or scope/recovery cross-tabulations requires linking the run-level review decisions to the original event records; aggregate counts alone do not provide that mapping.
Because the review covers only originally diagnosed runs, it does not estimate missed diagnoses among the remaining runs. Independent expert review is distinct from acceptance by the original developers or benchmark maintainers.

\section{Full Classification, Task Distribution, and Sensitivity}
\label{app:distribution}

\subsection{Primary Labels and Original Evaluation Outcomes}
\label{app:taxonomy}

\begin{table}[htbp]
\centering\small
\caption{Complete distribution of primary labels in the final audit. Terminal diagnoses include disputed-scope cases. Each trial receives one primary label; coexisting diagnoses and recovery events are recorded separately. ``No failure diagnosed'' denotes runs for which available evidence did not locate a specific failure.}
\label{tab:audit-taxonomy}
\begin{tabularx}{\linewidth}{Yrrr}
\toprule
Primary label & \shortstack[r]{Original outcome:\\passed} & \shortstack[r]{Original outcome:\\failed} & Total \\
\midrule
Terminal propagation diagnosis & $146$ & $327$ & $473$ \\
Propagation omission recovered within the trajectory & $133$ & $0$ & $133$ \\
Specification/interface mismatch & $39$ & $642$ & $681$ \\
Local algorithm or boundary error & $67$ & $418$ & $485$ \\
Compatibility regression; propagation chain not located & $9$ & $14$ & $23$ \\
Environment/build & $0$ & $47$ & $47$ \\
Evaluation process anomaly & $0$ & $48$ & $48$ \\
Incomplete/empty patch & $0$ & $12$ & $12$ \\
No failure diagnosed & $871$ & $0$ & $871$ \\
Unresolved & $21$ & $85$ & $106$ \\
\midrule
Total & $1{,}286$ & $1{,}593$ & $2{,}879$ \\
\bottomrule
\end{tabularx}
\end{table}

The $133$ runs whose primary label is recovered propagation are fewer than the $370$ runs containing recovery events:
the remaining runs may also have terminal or other diagnoses with higher priority.
``Evaluation-process anomaly'' includes build conflicts caused by overlaying hidden tests or problems with evaluation mapping;
the corresponding patch diagnoses are recorded separately.
``Unlocalized compatibility issue'' records an observed regression whose propagation relation remains unresolved.

\subsection{Explicit Scope Disputes and Evidence Types}
\label{app:aggregation}

\paragraph{From event records to aggregate statistics.}
For each diagnosed event $j$ in run $r$ whose terminal or recovery state is established, we record $L_{rj}\in\{\mathrm{terminal},\mathrm{recovered}\}$ alongside the original tuple $e=(\Delta,O,C,D)$.
For the terminal-event strata reported here, we retain evidence labels $E_{rj}$ and scope labels $S_{rj}\in\{\mathrm{undisputed},\mathrm{disputed}\}$.
Following supplementary execution support, every retained terminal event has $E_{rj}=\mathrm{exec}$; the former static classification is retained only in the historical provenance.
These labels support aggregation; the $89\%$ agreement statistic compares the four-part tuple $e$.
For the $R=2{,}879$ runs, define
\[
\begin{aligned}
T_r&=\{j:L_{rj}=\mathrm{terminal}\},\qquad A_r=\mathbf{1}[T_r\ne\varnothing],\\
X_r&=\mathbf{1}[\exists j\in T_r:E_{rj}=\mathrm{exec}],\\
U_r&=\mathbf{1}[T_r\ne\varnothing\ \land\ \forall j\in T_r:S_{rj}=\mathrm{undisputed}].
\end{aligned}
\]
Thus $\sum_r A_r=473$, $\sum_r X_r=473$, $\sum_r U_r=290$, and $\sum_r U_rX_r=290$.
Every retained terminal event has execution support, so $X_r=A_r$ and $\sum_r A_r(1-X_r)=0$; no diagnosed run remains static-only or evidence-unresolved.
The $A_r=0$ group comprises $2{,}222$ not-identified runs and $184$ terminal-status-unresolved runs, reported separately in Table~\ref{tab:audit-evidence-detail}; it records absence of an established terminal diagnosis, not verified absence of defects.
Recovery events with $L_{rj}=\mathrm{recovered}$ are counted separately; a run can contain both terminal and recovery events.

With task mapping $t(r)$, task coverage is
\[
p_{\mathrm{task}}=\frac{1}{64}\sum_{t=1}^{64}\mathbf{1}[\exists r:t(r)=t\ \land\ A_r=1]
=\frac{37}{64}=57.8\%.
\]
The event denominator is $N_{\mathrm{event}}=\sum_r|T_r|=546$, with $\sum_r\sum_{j\in T_r}\mathbf{1}[E_{rj}=\mathrm{exec}]=546$ execution-supported events and $0$ static-only events.
Their shares are $546/546=100\%$ and $0/546=0\%$, respectively.
The $85$ relation types are counted after deduplication within each task, as detailed in Table~\ref{tab:relations}.

\begin{table}[t]
\centering\small
\caption{Mutually exclusive run groups after manual scope review. Each run is grouped by the scope decisions for all of its terminal events; sources for the supplementary adjudication appear in the text.}
\label{tab:audit-sensitivity}
\begin{tabularx}{\linewidth}{Yrrr}
\toprule
Run group by scope & \shortstack[r]{Original outcome:\\passed} & \shortstack[r]{Original outcome:\\failed} & Total \\
\midrule
All terminal events undisputed & $75$ & $215$ & $290$ \\
Both undisputed and disputed events & $7$ & $19$ & $26$ \\
All terminal events disputed & $64$ & $93$ & $157$ \\
\midrule
Total & $146$ & $327$ & $473$ \\
\bottomrule
\end{tabularx}
\end{table}

The $473$ runs contain $546$ terminal events. Our supplementary manual scope review
classified all $334$ events that previously lacked a separate scope field as undisputed; together with the original $6$ undisputed events, this yields $340$ undisputed events,
while the remaining $206$ events have scope disputes. The submitted \path{labels.jsonl} records each terminal event's final scope status by event identifier.
We use these final manual scope decisions in the statistics.

Table~\ref{tab:audit-sensitivity} partitions runs into three mutually exclusive categories: $290$ runs with all terminal events undisputed,
$26$ with both undisputed and disputed events, and $157$ with all terminal events disputed.
The first two categories total $316$ runs and cover $340$ undisputed events; the other, disputed events in the mixed category are excluded from the undisputed counts.
All $206$ disputed events and all $340$ undisputed events now have execution support; scope disputes remain unchanged.

All $290$ runs whose terminal events are undisputed also have execution support: $75$ originally passed and $215$ failed.
These $290$ runs contain $314$ terminal events, all execution-supported.
The execution-support criterion therefore gives the same $290$-run subset as the undisputed-scope criterion, rather than a further nested group.
These nested counts can be recomputed from the event labels and run groups in the submitted \path{labels.jsonl}.

\paragraph{Stratification by evidence source.}
All $546$ terminal event diagnoses have undergone human review and now have execution support; static-only and evidence-unresolved event counts are both $0$.
The revised labels incorporate the authors' confirmation that supplementary execution support has been completed for all $367$ events formerly classified as source-only. Historical labels and source-level derivations are retained in the diagnostic provenance.
The original $179$ execution-supported events comprised $155$ test witnesses, $18$ in-trajectory probes, $5$ repair replays, and $1$ direct compiler diagnostic; this breakdown describes the original evidence, not the supplementary $367$ events.
Execution support requires a recorded failure linked to the same consumer relation in the final patch, and does not by itself imply a completed local repair replay. Scope status is recorded separately.

At the run level, all $473$ diagnosed runs have execution-supported terminal events, with $0$ static-only or evidence-unresolved runs.
Unresolved evidence here concerns the $473$ runs and is distinct from the $184$ runs with unresolved terminal status in the original full corpus.
We aggregate statistics under this common evidence stratification. The submitted labels provide the event-level classifications, evidence levels, and scope decisions.
These classifications incorporate the existing execution records, source code, trajectories, and author-confirmed supplementary execution support. The coverage of the original evidence supplied with this submission is specified in Appendix~\ref{app:artifact}; the reclassification does not expand the five-case repair-replay set.

\begin{table}[htbp]
\centering\small
\caption{Full evidence and terminal-status distribution after supplementary execution support. All $473$ diagnosed runs have execution-supported terminal events; static-only and evidence-unresolved counts are zero. Not identified and unresolved status retain their distinct diagnostic meanings.}
\label{tab:audit-evidence-detail}
\begin{tabularx}{\linewidth}{Yrrrrrr}
\toprule
\shortstack[l]{Original\\evaluation\\outcome} & \shortstack[r]{Any\\exec.} & \shortstack[r]{Static\\only} & \shortstack[r]{Unresolved\\evidence} & \shortstack[r]{Not\\identified} & \shortstack[r]{Unresolved\\status} & Total \\
\midrule
Passed & $146$ & $0$ & $0$ & $1{,}087$ & $53$ & $1{,}286$ \\
Failed & $327$ & $0$ & $0$ & $1{,}135$ & $131$ & $1{,}593$ \\
\midrule
Total & $473$ & $0$ & $0$ & $2{,}222$ & $184$ & $2{,}879$ \\
\bottomrule
\end{tabularx}
\end{table}

\subsection{Task Concentration}

\paragraph{Tasks, relations, and runs.}
Table~\ref{tab:relations} reports $37$ tasks, $85$ classes of within-task consumer relations, $473$ runs, and $546$ terminal events side by side.
Grouping uses the original diagnostic event as its unit: within a task, events with the same producer-side semantic change, obligation, and existing consumption path are grouped into one class,
without counting differences in field names, implementation variants, evidence levels, or input values as separate classes; distinct obligations or independent consumption paths remain separate.
When an original event already records a tightly coupled consumption chain or an associated set of consumers, that granularity is retained instead of splitting it into multiple call-graph edges;
each original event belongs to exactly one group, while specific consumer paths may overlap. The supplement records the group membership of each event.

The $146$ runs that passed the original evaluation cover $21$ tasks and $37$ relation classes; the $327$ failed runs cover $35$ tasks and $74$ relation classes.
The two pools share some tasks and relations, so the overall counts use their union.
This section describes the distribution of diagnosed terminal relation types and their run coverage.
The submitted \path{relations.jsonl} records relation identifiers, titles, consumer paths, member event and run identifiers, and counts.
\begin{table}[!htbp]
\centering\small
\caption{Within-task aggregation of relations in terminal diagnoses. Relation types are distinguished by the producer's semantic change, the adaptation obligation, and the existing consumer path; relations are not merged across tasks. A run may involve multiple relation types; runs per task are deduplicated using unique run identifiers.}
\label{tab:relations}
\begin{tabularx}{\linewidth}{Yrrr}
\toprule
Task identifier & \shortstack[r]{Relation\\types} & Runs & Events \\
\midrule
\path{abs-stepped-slices} & 1 & 10 & 10 \\
\path{anko-default-function-arguments} & 6 & 25 & 46 \\
\path{anko-typed-variable-bindings} & 3 & 12 & 12 \\
\path{arcane-drift-detection-baselines} & 1 & 21 & 21 \\
\path{arktype-json-schema-refs-depende} & 1 & 14 & 14 \\
\path{awilix-async-container-initializ} & 2 & 6 & 6 \\
\path{claude-code-by-agents-recursive} & 1 & 1 & 1 \\
\path{cliffy-config-file-parsing} & 1 & 4 & 4 \\
\path{drizzle-orm-window-function-buil} & 1 & 1 & 1 \\
\path{dynamodb-toolbox-conditional-att} & 3 & 11 & 14 \\
\path{dynamodb-toolbox-lazy-recursive} & 2 & 9 & 9 \\
\path{effect-sse-httpapi-streaming} & 1 & 1 & 1 \\
\path{expr-try-catch-errors} & 2 & 34 & 35 \\
\path{helm-array-merge-strategies} & 4 & 44 & 49 \\
\path{helm-unified-manifest-stream} & 3 & 6 & 7 \\
\path{ink-grid-box-layout} & 5 & 6 & 7 \\
\path{kea-atomic-signal-selectors} & 5 & 33 & 45 \\
\path{koota-composite-trait-aspects} & 4 & 14 & 14 \\
\path{koota-deferred-mutation-buffer} & 2 & 5 & 5 \\
\path{koota-pair-relation-tracking} & 3 & 10 & 11 \\
\path{koota-query-predicates} & 3 & 6 & 6 \\
\path{kysely-window-grouping-helpers} & 1 & 1 & 1 \\
\path{meriyah-explicit-resource-declar} & 2 & 2 & 2 \\
\path{ofetch-per-origin-circuit-breake} & 1 & 1 & 1 \\
\path{optique-conditional-option-depen} & 1 & 4 & 4 \\
\path{prometheus-transactional-reload} & 2 & 6 & 6 \\
\path{query-persist-restored-query-sta} & 1 & 1 & 1 \\
\path{quill-shared-toolbar-focus} & 2 & 22 & 25 \\
\path{scriggo-method-declarations} & 4 & 38 & 46 \\
\path{sql-formatter-bigquery-pipe-form} & 1 & 3 & 3 \\
\path{superjson-error-stack-serializat} & 2 & 8 & 8 \\
\path{tengo-callable-instance-isolatio} & 3 & 3 & 3 \\
\path{tengo-destructuring-bindings} & 5 & 40 & 57 \\
\path{termenv-preserve-ansi-resets} & 1 & 41 & 41 \\
\path{vitest-duration-sharding} & 1 & 2 & 2 \\
\path{yaegi-go-embed-directives} & 1 & 24 & 24 \\
\path{ytt-jsonpath-query-api} & 3 & 4 & 4 \\
\midrule
Total (37 tasks) & 85 & 473 & 546 \\
\bottomrule
\end{tabularx}
\end{table}

The five tasks with the most terminal annotations are Helm array merging with $44$ runs, termenv with $41$,
Tengo destructuring with $40$, Scriggo with $38$, and Expr with $34$,
totaling $197$ runs, or $41.6\%$ of all terminal annotations.
The five tasks with the most recovery records account for $151/370=40.8\%$.
We provide complete per-task counts linked to individual run diagnoses through stable task identifiers.

Of the $45$ runs for the Helm array-merging task, $44$ have terminal propagation annotations.
Among them, $43$ record the same repeated-concatenation mechanism, all now supported by supplementary execution evidence and retaining task-scope dispute flags.
These $43$ records count as one class in the relation summary; all $49$ terminal events for the task are grouped into $4$ relation classes.
Replay confirmation for the other run, \texttt{BRvPrLE}, concerns the annotation policy not being carried through to the configuration-reuse stage
(Section~\ref{sec:replay}); the repeated-concatenation relation receives supplementary execution support separately from that repair replay.

\subsection{Previously Explored Tasks and the Remaining Tasks}

Using the task list recorded during the earlier exploration,
we partition the fixed population into the $24$ tasks explored earlier and the remaining $40$ tasks.
Table~\ref{tab:review3-sensitivity} uses all runs in each group, including both successes and failures;
the table reports the group distributions separately from the final event-level scope classifications.

\begin{table}[t]
\centering\small
\caption{Descriptive grouping of historically explored tasks and the remaining tasks. Percentages use the number of runs in each group as the denominator, and metrics can overlap. The scope row retains the explicit flags in the historical snapshot; Table~\ref{tab:audit-sensitivity} reports the groups after supplementary manual adjudication.}
\label{tab:review3-sensitivity}
\begin{tabularx}{\linewidth}{Yrr}
\toprule
Metric & \shortstack[r]{Exploratory group\\(24 tasks)} & \shortstack[r]{Remaining group\\(40 tasks)} \\
\midrule
Runs & $1{,}079$ & $1{,}800$ \\
Original evaluation passed & $183$ (17.0\%) & $1{,}103$ (61.3\%) \\
Original evaluation failed & $896$ (83.0\%) & $697$ (38.7\%) \\
Terminal diagnoses & $268$ (24.8\%) & $205$ (11.4\%) \\
Containing recovery events & $144$ (13.3\%) & $226$ (12.6\%) \\
Unresolved primary label & $47$ (4.4\%) & $59$ (3.3\%) \\
Unresolved terminal status & $44$ (4.1\%) & $140$ (7.8\%) \\
Containing explicit terminal scope disputes & $108$ (10.0\%) & $75$ (4.2\%) \\
\bottomrule
\end{tabularx}
\end{table}

The original failure share is $83.0\%$ in the explored group and $38.7\%$ in the remaining group, indicating that the earlier analysis concentrated substantially on difficult tasks.
The shares with terminal annotations are $24.8\%$ and $11.4\%$, respectively, describing the two groups under the common final audit criteria.
The shares with unresolved primary labels are $4.4\%$ and $3.3\%$, respectively, whereas the shares with unresolved terminal propagation status are $4.1\%$ and $7.8\%$;
these two unresolved rates measure uncertainty in the primary category and terminal propagation status, respectively.
Explicit scope disputes mark boundaries of consumer obligations.
Both groups underwent full human re-review under the same criteria; this partition describes historical exploration and is reported separately from the additional independently sampled blinded audit.
The remaining $40$ tasks are not a previously sealed, unseen holdout set.
This table reports descriptive shares for the fixed corpus; relation aggregation captures repetition within tasks, while evidence sources and scope status are reported in separate strata.

\subsection{Evaluation Outcomes and Program Diagnosis}

Among the original $1{,}593$ failed runs,
$1{,}385$ did not pass all new tests but passed all existing tests,
$149$ did not pass all tests in either group,
and $59$ passed all new tests but did not pass all existing tests.
$213$ runs contain signals of missing tests, including $166$ in which every non-passing item is missing;
$129$ contain timeout signals. These latter two signals may overlap.
These counts describe test-result signals and depend on the F2P/P2P partition, missing reports, and parent--child aggregation;
semantic propagation events are diagnosed separately from specific program relations.

For example, a build gate can block multiple tests at once;
a timeout in an existing test may stem from the environment or from a hang introduced by the edit.
In some Obsidian records, both new and existing tests passed, but the evaluation mapping failed to recognize their names.
These records establish specific evaluation false negatives.
We retain the original scores and record diagnoses separately, linking them by run identifier for statistical analysis.

\section{Local Replay Setup and Validation Results}
\label{app:replaydetails}

For each of the five replays, we preserve the repository starting point, original model patch, fixed checks, local repair diff, and execution outputs.
The submitted core directory's \path{replays/register.json} links the five local replay kits to their run identifiers, diagnosed consumers, recorded results, and hashes.
The five replayed model patches are byte-for-byte identical to their archived counterparts, and the result-file hashes match the linked records.

The supplementary regression runs use the original DeepSWE task tests and designated images; the content hashes of all four task directories match the original run lockfiles.
Both versions use the same test lists, and new failures are compared individually by test identifier (Table~\ref{tab:original-regression}).
Table~\ref{tab:original-regression} and the case paragraphs below report the settings, repair versions, and results of these additional checks. The submitted replay kits correspond to their respective local-validation versions.

\begin{table}[htbp]
\centering\small
\caption{Supplementary reruns of the original DeepSWE evaluations. F2P reports passing counts before and after repair; P2P reports the passing counts shared by both versions. New failures are counted relative to the original model patch.}
\label{tab:original-regression}
\begin{tabular}{lccc}
\toprule
Task & F2P: before $\to$ after & P2P: both & New failures \\
\midrule
Kysely & $254/254\to254/254$ & $22/22$ & $0$ \\
Query & $7/8\to8/8$ & $42/42$ & $0$ \\
Helm & $44/47\to47/47$ & $10/12$ & $0$ \\
Expr & $73/79\to78/79$ & $66{,}265/66{,}265$ & $0$ \\
\bottomrule
\end{tabular}
\end{table}

\paragraph{Kysely: \texttt{wq82S5C}.}
The replay executes AST transformations and SQL generation on the actual source code using Node's TypeScript transformation support.
It checks two NULLS modifiers across three transformation paths and retains two direct-compilation controls.
In the submitted core directory, \path{cases/wq82S5C/verifier/test-stdout.txt.gz} records the verifier's cleanup and application of
\texttt{test.patch}; the corresponding \path{ctrf.json.gz} records $254$ passing F2P tests and $22$ passing P2P tests.
The supplementary rerun uses the original task tests and designated image: the original patch and the one-line repair both pass the same $276$ tests, with no new failures.
The fixed propagation checks yield $0/6\to6/6$ passes, and the direct-path controls remain $2/2\to2/2$.
Under Node~24.12.0 and TypeScript~5.9.3, \texttt{tsc --noEmit} passes for the production ESM and CJS code and the Node test code in both versions.
The semantic checks observe AST transformations and generated SQL; execution covers the tests selected by the original evaluation and the probes above.
The submitted \path{replays/kysely/} kit reproduces the AST/SQL checks. Its type-check log records the initial environment, while the successful supplementary checks use the environment specified above.

\paragraph{Query: \texttt{MYCfNHF}.}
The replay runner uses esbuild~0.27.3 from the original lockfile and the original tsconfig,
compiles $24$ actual core modules, and fixes a set of $12$ checks.
Four infinite-query targets cover the combinations of synchronous/asynchronous persisters and clients/observers;
eight controls cover ordinary restoration and new requests.
Adding a net four lines of unwrapping logic at the shared result-consumption point restores the targets.
The submitted \path{replays/query/} kit contains the fixed source, build dependencies, and local checks; an offline rerun in a separate directory has been completed.
The supplementary original evaluation improves from $49/50$ to $50/50$.
Runtime tests for the four packages---query-core, persist-client-core, React, and Preact---yield $1{,}502$ passes and $1$ pre-existing skip in both versions.
The repair includes an explicit \texttt{TData} type annotation at the shared consumption point and passes the full type-checking matrix for all four packages. The annotation leaves the generated JavaScript unchanged.
The submitted local kit uses \texttt{let data = ...}; the subsequent four-package validation uses \texttt{let data: TData = ...}. Both versions implement the same runtime unwrapping logic.

\paragraph{termenv: \texttt{5FGCqii}.}
The repair consumes the state in \texttt{style.go} and coordinates repeated style reopening in \texttt{ansi/truncate.go},
reusing the model's existing tokenizer.
Both versions use the same color-environment normalization and test expectations.
The original suite reports $122$ passing parent-test and subtest events, covering $76$ top-level tests.
The separate zero-width OSC truncation target fails in both versions.
Intermediate implementations and outputs are included in the core directory's \path{replays/termenv/development_initial/}.

\paragraph{Helm: \texttt{BRvPrLE}.}
The replay calls \texttt{Upgrade.reuseValues} in the actual action package,
using the existing test infrastructure without requiring a Kubernetes cluster.
The repair changes two source files, adding $17$ lines and deleting $2$:
it adds a thin wrapper around the existing annotation parser and passes the policy to the configuration-reuse consumption stage while preserving CLI precedence.
The $48$ checks that remain passing in the table comprise $9$ local controls,
$5$ existing action regression leaf tests, and $34$ util leaf tests.
In the supplementary original-evaluation rerun, three upgrade failures are resolved, two original lint failures remain, and no new failures occur.
After retaining three fixture symbolic links in the original task image, additional tests cover five packages: action, util, two loaders, and the filesystem.
Both versions pass $374/374$ deduplicated leaf tests, including six permission checks rerun as a non-root user.
The submitted \path{replays/helm/} kit covers the five local targets and the $48$ preserved checks; the additional original-evaluation and $374$-test results are reported here.
The full audit also includes a distinct duplicate-merge diagnosis with supplementary execution support; this replay concerns the annotation-source relation.

\paragraph{Expr: \texttt{7uAWePB}.}
The original patch bound an error pointer to the catch variable, but the existing \texttt{string} path dereferenced it and lost access to its \texttt{Error()} method.
With the original patch fixed, \texttt{string(e)} produces a struct representation enclosed in braces.
After string conversion preserves the error value and uses its message semantics, all six targets and nine controls pass.
The original repository suite yields $66{,}786$ passing events and $1$ skip both before and after repair;
the suite counts parent-test and subtest events.
The repair adds a net eight lines and also makes \texttt{string(E)} use the error-message representation for error values supplied through the environment,
changing the output from \texttt{\{env err\}} to \texttt{env err}.
In the supplementary original-evaluation rerun, P2P remains at $66{,}265/66{,}265$ and F2P increases from $73/79$ to $78/79$, with no new failures.
Using the unmodified task source as the behavioral baseline, additional compatibility checks preserve $18/22$ existing behaviors.
All four changes involve non-nil pointer error values supplied through the environment switching to the message representation; an additional $10/10$ catch-method and type controls remain passing.
Thus, the repair restores the targets while extending some existing string-conversion semantics. The submitted \path{replays/expr/} kit contains the local checks and original repository-suite outputs; the additional compatibility and original-evaluation results are reported here.
The original evaluation still has a failing \texttt{ErrtypeTypeError} test;
the original evaluation and the six local targets are reported separately according to their check definitions.

\section{Positive and Negative Cases of Program Attribution and Process Evidence}
\label{app:boundaries}

\paragraph{Obligations for optional fields.}
A field that is optional in the type system may still carry semantics specified by a particular invocation.
In Kysely, the input contains a NULLS modifier, which the consumer discards, thereby changing the output.
This input--processing--output chain addresses path consistency more directly than type-checking results do.

\paragraph{Specifications and local implementations.}
If a requirement calls for a multiline report but the patch outputs a single line, the primary issue is specification implementation;
an off-by-one index boundary or acceptance of an invalid numeric format is primarily a local algorithm error.
We diagnose propagation by locating a distinct existing consumer's unmet adaptation obligation introduced by the agent's edit.

\paragraph{Transient inconsistency and recovery.}
Changing a definition before updating its callers is a normal editing sequence.
Our audit counts recovery only when an existing consumer exhibits an observable runtime behavioral failure and is subsequently repaired.
Compilation errors during an ongoing migration and synchronization of golden files do not count as such recovery;
unresolved terminal compilation problems may instead be assessed through their specific propagation relations.
These criteria address process events and final program states, respectively.

\paragraph{Information presentation and explicit judgments.}
Assistant steps are indexed from $1$, serialized message positions from $0$, and patch line numbers from $1$.
Visibility is determined by the tool-message content presented at the time; archived \texttt{raw\_output} may retain middle sections omitted from the displayed tool output.
The complete output supports program diagnosis; evidence of code presentation comes from the displayed messages.
Step $17$ of Kysely \texttt{5WQPAYr} supports code presentation,
and step $135$ of Helm \texttt{BRvPrLE} supports explicit discussion;
these provide message-level process evidence.
Step $130$ of Query \texttt{MYCfNHF} corresponds to message $260$ and records the inference based on a shared outer entry point.
Step $4$ of termenv \texttt{5FGCqii} corresponds to the read request in message $8$, while message $11$ presents the old renderer;
steps $32$ and $56$ correspond to messages $72$ and $120$, respectively, whose archived reasoning text asserts that \texttt{Styled} need not consume the state.
These locations separately establish visible code and explicit judgments, distinguishing a tool request from its displayed response.

\paragraph{Evidence levels for additional cases.}
Quill \texttt{JMoRtHX} retains the previous editor's image-input restrictions when multiple editors share a toolbar.
In Koota \texttt{BRMDKd8}, the edit stores \texttt{Not(aspect)} constraints in \texttt{forbiddenAspects}.
The ordinary query evaluator handles this representation, but the existing \texttt{checkQueryTracking} omits the constraint when a tracking modifier is combined with \texttt{Not(aspect)}.
This diagnosis now has supplementary execution support, with task scope still recorded as disputed.
These cases concern preservation of associated state and adaptation of parallel query-evaluation paths, respectively.
We also consider alternative process explanations involving code visibility, interface-design choices, and differences in specification interpretation.
We report these cases according to their recorded evidence sources.

\subsection{Supplementary Process Review}
\label{app:process-review}

Table~\ref{tab:original-condition-process} groups the frozen cohort by its original index condition; terminal and recovery events can coexist.
These are descriptive counts, separate from the dependency-review intervention, and do not identify an indexing effect.

\begin{table}[htbp]
\caption{Original index conditions and process outcomes. Counts use all runs in each condition; terminal and recovery columns overlap.}
\label{tab:original-condition-process}
\centering\small
\begin{tabular}{lrrrr}
\toprule
Condition & Runs & Terminal & Recovery & Both \\
\midrule
\texttt{bare} & 960 & 154 (16.0\%) & 138 (14.4\%) & 31 \\
\texttt{inject-file} & 959 & 170 (17.7\%) & 118 (12.3\%) & 30 \\
\texttt{guided} & 960 & 149 (15.5\%) & 114 (11.9\%) & 22 \\
\bottomrule
\end{tabular}
\end{table}

The process review selected six tasks with seed $20261110$ from the $24$ tasks containing both terminal and recovery events, then presented one terminal and one recovery case per task ($12$ events from $12$ runs).
Event selection is descriptive; recovery cases require complete public failure--repair--verification records. Within the same Kea run, the health-forwarding event replaced the earlier event in this case set. The retained case and its source references are in \path{kea_selector_health.json}.
Consumer code or contracts were confirmed as presented before the relevant edit in $10/12$ cases: $5/6$ terminal and $5/6$ recovery cases.
All six selected recoveries have recorded failure--repair--verification chains: three existing-test triggers, two agent-constructed probes, and one agent-added test.
Visibility uses actual displayed tool messages, excluding elided archival output; it does not establish understanding or later reuse of the information.
The selected cases neither estimate corpus-wide process frequencies nor explain the different outcomes of the two groups; this review is not an additional human blinded audit.
The process directory supplies the case annotations and $92$ source-linked quotations in \path{annotations.jsonl}, the corresponding public steps in \path{visible/}, and aggregate counts in \path{counts.json}.

Table~\ref{tab:process-evidence-chain} organizes these cases into a common process chain. This supplementary check uses public assistant content, issued commands, and displayed tool responses, distinguishing explicit statements from targeted editing actions while preserving the original event identities and labels.
Active retrieval requires a command targeting the selected consumer and actual presentation of the relevant code or contract; a filename or an unrelated excerpt from the same file is insufficient. Later reuse requires paired explicit records about the same obligation. Not established does not mean that reading, judgment, or reuse did not occur.

\begin{table}[htbp]
\centering\footnotesize
\caption{Evidence chains for the descriptive cases. T/R denote terminal/recovered status; C confirms code or contract presentation, and NE means not established. e marks the edit step; read positions use step:request/response message. J/A denote public judgment text/targeted edit, located by step/message. Chain numbers are assistant steps.}
\label{tab:process-evidence-chain}
\begin{tabularx}{\linewidth}{>{\raggedright\arraybackslash}p{.15\linewidth}>{\raggedright\arraybackslash}p{.13\linewidth}>{\raggedright\arraybackslash}p{.10\linewidth}>{\raggedright\arraybackslash}p{.16\linewidth}>{\raggedright\arraybackslash}p{.08\linewidth}Y}
\toprule
Case / run & Edit / read & Shown\newline pre/post & Judgment / action & Later reuse & Detection $\to$ repair $\to$ verification \\
\midrule
01-R\newline\texttt{NtJTSTr} & e129\newline20:40/41 & C/C & J: NE\newline A: 133/266 & NE & Probe 131 $\to$ C133 $\to$ suite 136 \\
01-T\newline\texttt{ftB5YxJ} & e29\newline5:10/11 & C/NE & NE & NE & Not established \\
02-R\newline\texttt{X7WyyaV} & e20\newline14:28/29 & C/NE & J: NE\newline A: 35/70 & NE & New test 31 $\to$ C35 $\to$ target 37 \\
02-T\newline\texttt{tZ5A8ae} & e37\newline4:8/9 & C/C & NE & NE & Not established \\
03-R\newline\texttt{uFXJqjF} & e90\newline100:201/202 & NE/C & J: NE\newline A: 102/205 & NE & Existing test 98 $\to$ P102 $\to$ suite 104 \\
03-T\newline\texttt{Kh8cqmU} & e80\newline73:146/147 & C/NE & NE & NE & Not established \\
04-R\newline\texttt{tyrMzUZ} & e83\newline18:36/37 & C/C & J: NE\newline A: 149/298 & NE & Probe 147 $\to$ C149 $\to$ target 150 \\
04-T\newline\texttt{XehcNMc} & e86\newline18:36/37 & C/NE & NE & NE & Not established \\
05-R\newline\texttt{RXC67jp} & e22\newline6:12/13 & C/C & J: 28/56\newline A: 28/56 & NE & Existing test 27 $\to$ P30 $\to$ suite 30 \\
05-T\newline\texttt{SwfxaaT} & e18\newline6:15/16 & C/NE & NE & NE & Not established \\
06-R\newline\texttt{jwZzGCn} & e41\newline9:22/24 & C/C & J: NE\newline A: 67/146 & NE & Existing test 64 $\to$ P67 $\to$ target 70 \\
06-T\newline\texttt{qpCWwvn} & e44\newline NE & NE/NE & NE & NE & Not established \\
\bottomrule
\end{tabularx}
\par\smallskip\raggedright\footnotesize
Original records are in the process directory's \path{visible/<sample_id>.jsonl}. The read column gives one verified request, not an exhaustive reading history; retrieval may occur after the edit, as in 03-R. C/P in repair chains denote consumer/producer locations.
01-R also has a recursion repair at step 135; 06-R adds state save/restore at step 68. Subsequent checks are not attributed to the preceding edit alone. For 06-R, the root-package targets recover while parser tests still fail to compile.
06-T reads the REPL portion of the same file without displaying the selected \texttt{addPrints} consumer. A records editing action, not an explicit verbal judgment or a revision of the original annotation.
\end{table}

\subsection{Supplementary Dependency-Judgment Sequences}
\label{app:prior-judgment}

In eicrud \texttt{zNSeKSU}, step $21$ actively reads the existing \texttt{\_doLimitQuery} consumer.
Before adding the server constraint, the archived reasoning at step $22$ concludes, ``we may not need to modify the client for this issue.''
Step $62$ adds a server rejection of requests combining \texttt{cursor} with \texttt{offset}.
The archived reasoning at step $128$ retains the client decision and states that ``the client would pass it through, and the server handles it. Good.''
The unchanged automatic-pagination consumer preserves \texttt{cursor} in the original options and adds \texttt{offset} when fetching another batch, conflicting with the new constraint.
This sequence links an earlier decision, a code change, and a later judgment retaining that decision.
The client-adaptation obligation is scope-contested and this interaction is statically derived in the existing audit; the case is supplementary and is not evidence for the strict undisputed-scope counts or an isolated cognitive cause.
The original trajectory is \path{deepswe_round2_trajectory_evidence/zNSeKSU.json.gz}; message indices $44$, $124$, and $256$ locate these judgments and the edit, and message $43$ displays the consumer.

Table~\ref{tab:focal-judgment-sequences} contrasts this sequence with the focal termenv judgment chain. These are separately selected process cases, outside the twelve-case group in Table~\ref{tab:process-evidence-chain}.
Source inspection places the quoted judgments below in archived assistant messages' \texttt{reasoning\_content} field, with empty corresponding \texttt{content}; read commands and code presentation are supported by tool requests and responses. Archived reasoning text, public assistant content, and tool actions are recorded separately. Archival presence alone does not establish verbatim inclusion in every subsequent request or identify an internal cognitive cause.

\begin{table}[htbp]
\centering\small
\caption{Dependency-judgment sequences in independent focal cases. s/m denote assistant step/archived message index. Judgment quotations come from \texttt{reasoning\_content} and are separate from the public-assistant-content coding.}
\label{tab:focal-judgment-sequences}
\begin{tabularx}{\linewidth}{>{\raggedright\arraybackslash}p{.16\linewidth}YYY}
\toprule
Case & Code presentation and edit & Explicit judgment sequence & Program consequence and boundary \\
\midrule
termenv\newline\texttt{5FGCqii} & s4/m8 requests \texttt{cat style.go}; m11 displays the old renderer. s22--23 introduce the state. & s32/m72: ``The preserveResets field only affects Truncate, not Styled.'' s56/m120: ``Only Truncate uses preserveResets. That's fine''. & Ordinary rendering still omits the new state. The explicit judgment is formed after the edit and persists during later review. \\
eicrud\newline\texttt{zNSeKSU} & s21/m42 reads the client; m43 displays the consumer. s62/m124 adds the server's \texttt{cursor}/\texttt{offset} restriction. & s22/m44: ``we may not need to modify the client for this issue''; s128/m256: ``the client would pass it through, and the server handles it. Good.'' & The tentative pre-edit decision remains after a new constraint. Client adaptation is scope-contested; the conflict is statically derived. \\
\bottomrule
\end{tabularx}
\par\smallskip\raggedright\footnotesize
The termenv source is \path{cases/5FGCqii/visible_messages.json} in the core directory; the eicrud source is \path{zNSeKSU.json.gz} in the supplementary trajectory directory. Directory mappings appear in Appendix~\ref{app:artifact}.
\end{table}

\subsection{Dependency-Review Prompt Intervention}
\label{app:dependency-review}

The intervention uses the same DeepSWE tasks and diagnostic rules, with $2{,}879$ runs in each group.
The normal-continuation baseline consists of the corresponding archived runs.
General checking and dependency review resume from the corresponding checkpoint in each original run, holding fixed the context and code state at resumption and matching the token-budget cap.
The intervention varies the review prompt inserted after each code edit: general checking requests implementation-correctness and test review; targeted review requests re-examination of dependencies and adaptation obligations.
The outcome uses the same undisputed-scope terminal-omission criterion across groups; the $290$ baseline runs correspond to the strict subset in the original audit.

\begin{table}[htbp]
\caption{Terminal omissions under the same undisputed-scope criterion. Each group contains $2{,}879$ runs.}
\label{tab:dependency-review}
\centering\small
\begin{tabular}{lrr}
\toprule
Group & Runs with terminal omissions & Share \\
\midrule
Normal continuation (archived) & 290 & 10.1\% \\
General correctness/test check & 276 & 9.6\% \\
Dependency-review prompt & 69 & 2.4\% \\
\bottomrule
\end{tabular}
\end{table}

The dependency-review prompt is:
\begin{quote}\small
Based on the code before and after the edit in the current context, check which existing dependency relationships or prior judgments have changed as a result of this edit. Review the changes in dependency relationships one by one.
\end{quote}

Dependency review yields $221$ fewer terminal-omission runs than the archived baseline (a $76.2\%$ relative reduction) and $207$ fewer than general checking ($75.0\%$).
The dependency-review group also achieves an overall task success rate 1 percentage point higher than the archived normal-continuation baseline.
These results support the practical value of explicitly re-deriving dependency obligations from the starting information available to the agent.
The trajectory evidence motivates the process explanation that retained judgments can remain premises for incorrect later reasoning; the intervention alone does not establish that prior-context trust caused every omission or isolate the contributions of reasoning capacity and workflow constraints.

\subsection{Natural Recovery in Kysely}
\label{app:kysely-recovery}

\begin{figure}[t]
\centering
\includegraphics[width=\linewidth]{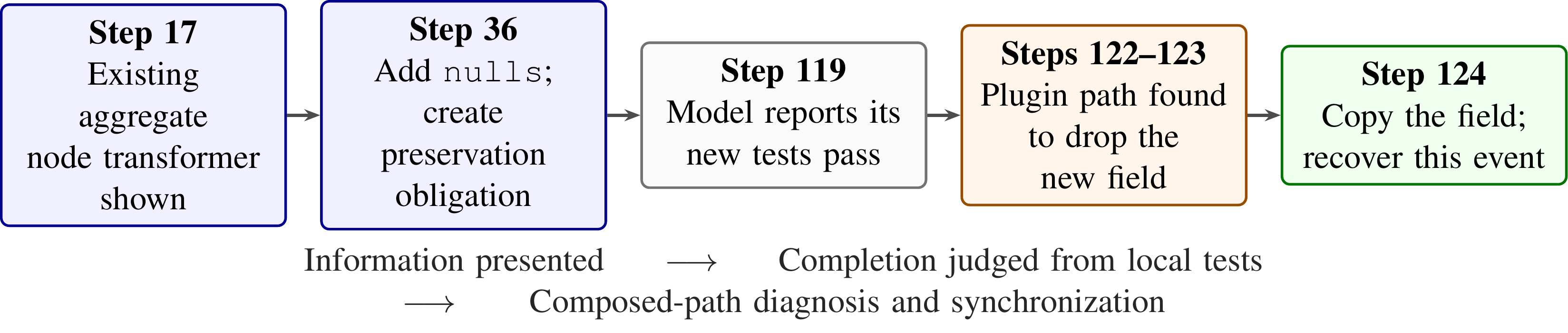}
\caption{Natural recovery in Kysely \texttt{5WQPAYr}. Steps are assistant steps numbered from $1$;
horizontal spacing indicates order only. Step $119$ records the model's statement about its own tests; subsequent actual diagnosis localized the gap in the plugin path.
This run is independent of the terminal record in Figure~\ref{fig:propagation}.}
\label{fig:recovery}
\end{figure}

On the Kysely NULLS-modifier task, the independent \texttt{inject-file} run \texttt{5WQPAYr} exhibits a traceable natural recovery process (Figure~\ref{fig:recovery}).
The old aggregate transformation function was displayed in full before the new field was added.
The model later reported that its newly added tests passed; only afterward did it discover that plugin transformation dropped the field and add the missing copy.
This independent run provides the complete temporal sequence of code reading, modification, gap discovery, and recovery through synchronization.

\section{Reviewable Artifacts and Source Mapping}
\label{app:artifact}

\paragraph{Submitted directories and coverage.}
The submitted data supplement contains five top-level directories:
\path{deepswe_core_evidence_slim} (\textbf{C} in Table~\ref{tab:artifact}),
\path{deepswe_473_terminal_evidence} (\textbf{D}),
\path{deepswe_process_audit_20261110} (\textbf{P}),
\path{deepswe_round2_trajectory_evidence} (\textbf{R}), and
\path{codex_72_baseline_model_responses} (\textbf{B}).
Directory C contains all $2{,}879$ original scores and final patches, $64$ task descriptions, a minimal frozen-label projection, nine selected visible interaction records with their archived verifier reports, and five local replay kits.
Directory D contains labels for $473$ runs and $546$ terminal events, together with membership and counts for $85$ within-task relations.
Directory P contains public-step projections and annotations for 12 events across six tasks, including 92 source-linked quotations.
Directory R provides the eicrud trajectory and an index with its checksum and message coordinates for Appendix~\ref{app:prior-judgment}.
Directory B contains the records, evaluation logs, and diagnosis labels for the $72$ Codex baselines in Appendix~\ref{app:codex-featurebench}.

\paragraph{Reading the current labels.}
In D, \path{labels.csv} has one row per terminal event, and \path{labels.jsonl} groups events by run. Both record evidence level, scope status, and relation identifiers.
\path{relations.jsonl} provides relation membership and counts.
Use the current labels and relation memberships for Tables~\ref{tab:audit}, \ref{tab:audit-sensitivity}, and \ref{tab:relations}; C contains the earlier frozen labels.

\begin{table}[htbp]
\caption{File index for the data supplement. Paths are relative to the indicated directory.}
\label{tab:artifact}
\centering\small
\begin{tabularx}{\linewidth}{>{\raggedright\arraybackslash}p{.25\linewidth}cY}
\toprule
Content & Directory & Relative path \\
\midrule
Original scores and patches & C & \path{data/scores.csv}; \path{patches/} \\
Task descriptions & C & \path{tasks/} \\
Frozen labels and census & C & \path{data/final_labels.jsonl}; \path{data/expected_counts.json} \\
Nine selected interactions and verifier reports & C & \path{cases/index.json}; \path{cases/<trial>/visible_messages.json}; \path{cases/<trial>/verifier/} \\
Five local replay kits & C & \path{replays/register.json}; \path{replays/} under \texttt{kysely}, \texttt{query}, \texttt{termenv}, \texttt{helm}, \texttt{expr} \\
Current terminal labels & D & \path{labels.csv}; \path{labels.jsonl}; \path{counts.json} \\
Relation membership & D & \path{relations.jsonl} \\
Process annotations and public steps & P & \path{annotations.jsonl}; \path{annotations.csv}; \path{visible/}; \path{counts.json} \\
Supplementary judgment sequence & R & \path{zNSeKSU.json.gz}; \path{index.json} \\
Codex baseline records and labels & B & \path{index.csv}; \path{records/}; \path{diagnoses.jsonl}; \path{counts.json} \\
File checksums & Root & \path{SHA256SUMS} \\
\bottomrule
\end{tabularx}
\end{table}

\paragraph{Verification and evidence navigation.}
From the supplement root, \texttt{sha256sum -c SHA256SUMS} checks the distributed files against the listed hashes.
Trial identifiers join C's scores and frozen labels to the patches and D's current terminal labels. Event identifiers link D's labels to its relation membership.
Use C's \path{cases/index.json} for its nine interaction records and P's \path{annotations.jsonl} for the 12 process cases. For each process case, its sample identifier locates the corresponding file in \path{visible/}, and the annotation gives message coordinates for the quotations.
C's \path{replays/register.json} identifies the five local replay kits and their recorded results.
The corresponding appendices report the settings, results, and check scopes of the additional original-evaluation, type-checking, extended-regression, SWE-bench Verified, and FeatureBench analyses. Table~\ref{tab:artifact} lists the files included in this supplement.

\section{Failure Diagnosis and Program-Level Counterexample Validation on SWE-bench Verified}
\label{app:sweverified}

Using our saved evaluation records, this section combines instance-level diagnosis of the subset that failed the original evaluation with replays of three successful patches
to examine failure categories, recovery during the trajectory, and evaluation blind spots in another benchmark and coding framework.
Archive verification establishes the correspondence among instances, configurations, patches, and evaluation outcomes; program replays identify specific propagation gaps and verify the effects of local synchronization.

\subsection{Dataset, Configurations, and Evaluation Correspondence}

The extended dataset comes from \path{baseline-full-trajectories.tar.xz} and contains one final saved trajectory for each of $500$ instances,
with $30{,}925$ steps across $12$ repositories and $499$ distinct repository--base-commit combinations.
Instance-by-instance comparison with a pinned version of the official Verified dataset confirms exact agreement in instances, repositories, base commits, and archived original problem statements.\footnote{Official dataset: \url{https://huggingface.co/datasets/SWE-bench/SWE-bench_Verified}. The dataset snapshot and file hashes identify the version used in the study archive.}
The original archive preserves message histories, tool observations, and inputs at each step as the source trajectory records; a more readable copy omits duplicate nested copies of inputs.

All runs use \texttt{deepseek/deepseek-v4-flash}, SWE-agent~1.1.0, and SWE-ReX~1.4.0,
with temperature $0$ and top-p $1$.
For $411$ runs, the call limit is $100$ and the configured cost limit is $3$;
for $88$ runs, these values are $100$ and $0$, respectively, and for the remaining $1$ run, they are $200$ and $3$; $0$ is reported solely as the original configuration value.
The script permits retrying empty patches with a higher call limit and overwriting earlier outputs, so these $500$ trajectories are the final saved trajectories, including retry results.
The ratio $462/500$ is the archived pass rate for these trajectories, which differs in its counting protocol from first-attempt \texttt{pass@1} under a uniform budget.
Some trajectories also consulted upstream issues, PRs, or newer implementations, with external repair materials accessible during task solving.

Both the per-instance \path{report.json} files and the aggregate table record $462$ passing and $38$ failing instances.
All patches are nonempty and were applied successfully; \texttt{pylint-dev\_\_pylint-4604}
also carries an ambiguous \texttt{no\_tests\_collected} flag, so its failed evaluation status retains an explanation at the test-collection level.
For every instance, the final patch in the trajectory, the prediction file, and \path{patch.diff} in the evaluation directory have identical SHA-256 hashes.
We obtained these counts by executing the official evaluation and verified them against the saved per-instance reports; instance-level failure diagnoses and the three local replays below are reported separately.

\subsection{The 38 Runs That Failed the Original Evaluation}
\label{app:sweverified-failures}

We fix the diagnostic scope to the $38$ runs marked \texttt{resolved=false} in the original archive,
checking each problem statement, complete final patch, original evaluation log, and test patch, and tracing edits, failures, and subsequent corrections through the relevant trajectories.
We reviewed the consumer obligations and classification boundaries and retained the supporting evidence.
All $38$ records and their evidence-based diagnoses have been individually checked and verified by human reviewers; Table~\ref{tab:sweverified-failures} summarizes the results.
The primary diagnosis explains why the original run ultimately failed the evaluation; recovery events are recorded separately.

\begin{table}[htbp]
\centering\small
\caption{Primary diagnoses for the 38 SWE-bench Verified runs that failed the original evaluation. Each run receives one terminal primary diagnosis; two recovery events are counted separately from the primary diagnoses.}
\label{tab:sweverified-failures}
\begin{tabularx}{\linewidth}{YrY}
\toprule
Primary diagnosis & Runs & Evidence \\
\midrule
Terminal propagation omission (human-confirmed) & $2$ & Sphinx~9461 passes all $62$ original evaluation tests after local synchronization; Astropy~13398's mismatch between new and existing coordinate-transformation paths is human-confirmed \\
Incomplete repair & $17$ & Branches, inputs, or call paths required by the original task remain uncovered, including Django~11734 \\
Specification/interface or representation mismatch & $11$ & Implementation outputs differ from acceptance requirements, including strict representation requirements \\
Local algorithmic error & $7$ & Algorithmic or conditional errors identified, without an independently established propagation relation \\
Evaluation-process anomaly & $1$ & pylint~4604 fails during import; no tests are actually executed \\
\midrule
Total & $38$ & Originally failing subset, retaining original evaluation outcomes \\
\bottomrule
\end{tabularx}
\end{table}

\paragraph{Confirmed terminal propagation omission.}
In Sphinx~9461, autodoc newly generates the \texttt{:classmethod:} option,
but the existing \texttt{PyProperty} parser has not been adapted; the original log explicitly reports \texttt{unknown option: classmethod}.
On the pinned source with the complete original model patch applied, adding only three lines for the consumer's option definition and signature-prefix handling
changes the same original evaluation suite from $61$ passes and $1$ failure to $62$ passes;
the corresponding unmodified source yields $59$ passes and $3$ failures.
An actual HTML build also verifies restoration of documentation, types, registration, and references for ordinary properties, abstract properties, class properties, and abstract class properties.
The RST generated by autodoc is identical before and after the repair, preserving the added functionality and original assertions.
This replay confirms the consumer relation under the listed checks.
This case is reported separately from the three program-level counterexamples in the next subsection that passed the original evaluation.

Astropy~13398's new coordinate-transformation route produces numerical discrepancies in existing route-equivalence tests.
Human review confirmed this relation as a terminal propagation omission based on the original execution records and review of the consumer obligation;
Sphinx~9461 additionally has a local replay comparison before and after consumer synchronization.

\paragraph{Two recovery events.}
In Django~11087, narrowing the fields in deletion queries causes the existing \texttt{NestedObjects} display path to issue an additional deferred-loading query,
increasing the original two queries to three; synchronizing field selection restores the count to two.
This recovery in performance behavior is recorded separately from the final local algorithmic issue of still selecting the extra \texttt{origin\_id} field.
In scikit-learn~25102, passing the new \texttt{dtypes} argument to an existing override of \texttt{set\_output}
triggers a \texttt{TypeError}; maintaining compatibility with the old call restores the same test group.
This approach is later withdrawn, and the final type-preservation requirement remains unmet.
The two runs receive primary diagnoses of local algorithmic error and incomplete repair, respectively, with recovery events recorded separately.

\paragraph{Classification and evaluation boundaries.}
Additional validation of Django~11734 shows that the original model patch combined with the consumer repair restores the original query test group,
but another explicit nested-reference path in the unmodified source also exhibits the same kind of stringification.
The boundary between a new consumer obligation and an incompletely repaired pre-existing consumer defect is insufficiently clear, so we conservatively classify this run as an incomplete repair and retain it in the set.
In Django~16263, a local optimization algorithm error in annotation pruning breaks count/grouping equivalence.
Removing and then restoring the \texttt{None} guard in the same function in pylint~4604 is an ordinary local correction;
its final evaluation fails to collect tests because the test code imports the nonexistent \texttt{IS\_PYPY}.
A local check of SymPy~21596 produces \texttt{FiniteSet(1/2)},
whereas the original evaluation requires a mathematically equivalent unevaluated set; this case is reported as a difference in representation requirements.
Some runs' evaluation reports contain truncated test identifiers or differences in assigned names; diagnosis uses the actual failure logs for localization, and original evaluation outcomes are retained.
This table uses the fixed historical subset of $38$ originally failing runs; subsequent problem-statement corrections and reevaluations (see Section~\ref{app:statement-correction}) are recorded separately.

\subsection{Three Program-Level Counterexamples That Passed the Original Evaluation}

For each of the three cases, we apply the complete original model patch to its pinned starting revision and then compare it with a locally synchronized version;
program inputs and expected correctness remain consistent with the original task's repair requirements.
The original evaluation columns count test identifiers in our saved official evaluation reports; the three cases are purposefully selected program-level counterexamples for validation through execution.

\begin{table}[htbp]
\centering\small
\caption{Three supplementary SWE-bench Verified replays. All passed the original evaluation; synchronization repairs confirm only the listed consumer relations.}
\label{tab:sweverified-replays}
\begin{tabularx}{\linewidth}{>{\raggedright\arraybackslash}p{.15\linewidth}>{\raggedright\arraybackslash}p{.14\linewidth}YY}
\toprule
Instance & Original F2P/P2P & Diagnosis with the complete model patch & After local synchronization \\
\midrule
Sphinx 7440 & $1/1$; $12/12$ & \texttt{:any:`MySQL`} cannot find the case-preserving glossary entry & Reference and HTML restored; the original repair allowing differently cased entries to coexist is preserved \\
SymPy 12419 & $1/1$; $25/25$ & For a range-constrained delta in a product, reduction before substitution yields $1$, whereas direct substitution yields $0$ & Both orders yield $0$; controls for the old two-argument form, in-range cases, empty products, and the original task are preserved \\
xarray 6461 & $1/1$; $247/247$ & \texttt{where(..., keep\_attrs=True)} overwrites coordinate units in seconds with the variable's kelvin units & Coordinate attributes restored; the original task's scalar-input repair is preserved \\
\bottomrule
\end{tabularx}
\end{table}

\paragraph{Sphinx: glossary keys and the general-reference consumer.}
The edit in \texttt{sphinx-doc\_\_sphinx-7440} preserves the case of glossary keys
and adapts the dedicated \texttt{:term:} path, but misses the old \texttt{resolve\_any\_xref} lookup, which lowercases glossary entries uniformly.
Building a minimal HTML document with the actual source shows that the previously working \texttt{:any:`MySQL`} reference consequently fails to resolve.
Reusing the adapted glossary resolver in the glossary branch of general-reference resolution restores the reference.
The original functionality fixture requires differently cased entries to coexist: strict-build exit codes at the original starting revision, with the model patch, and with the synchronized version are
$1\to0\to0$; for the consumer-path fixture, they are $0\to1\to0$.
The restored reference fixture's HTML is byte-for-byte identical to its pre-edit output, as verified through actual Sphinx builds.

\paragraph{SymPy: a new range constraint and existing product reduction.}
\texttt{sympy\_\_sympy-12419} adds an optional range to \texttt{KroneckerDelta},
allowing sums over identity-matrix elements to use dimension bounds; the old \texttt{\_remove\_multiple\_delta}
and \texttt{\_simplify\_delta} still reconstruct two-argument objects, discarding the range.
Let $d=\operatorname{KroneckerDelta}(i,j,(0,2))$ and
$P=\operatorname{Product}(d,(i,a,b))$, then substitute $a=b=j=5$.
With the original patch, reducing $P$ before substitution yields $1$, whereas substituting before evaluating the product yields the correct value $0$.
This input uses a finite-range delta accepted by the new interface, and its construction requires no out-of-bounds indexing into an identity matrix.
After the two consumers conservatively retain expressions for range-constrained objects, both paths yield $0$;
controls for the old two-argument form, in-range equality/inequality, empty products, and the original task's identity-matrix summation are all preserved.
The original starting revision supports only two arguments, so this case tests semantic preservation when a new representation enters existing consumers.
The conservative repair aims to preserve range semantics and may reduce symbolic simplification; other local errors in the new API remain outside the scope of the repair.
All $29$ existing local delta and tensor-function tests pass at each of the three stages; the original model patch passes all $1+25$ tests in the official evaluation we executed.
The execution environment uses Python~3.9.25 and mpmath~1.1.0.

\paragraph{xarray: the scope of consumption for variable and coordinate attributes.}
To repair scalar inputs, \texttt{pydata\_\_xarray-6461} changes the attribute callback to capture the top-level variable's \texttt{x.attrs}.
The existing coordinate-construction and merge paths reuse the same callback, thereby writing the variable's \texttt{units=K} into a time coordinate that originally has
\texttt{units=s}, while losing the coordinate's own \texttt{axis=time}.
The replay uses actual \texttt{xr.where} calls to construct arrays with variable and coordinate attributes;
preserving and restoring the corresponding attributes by coordinate name restores the target coordinate's original values while preserving numerical results and the scalar-input repair.
Controls with no attributes, variable attributes only, and coordinate attributes check numerical values and attributes separately.
The execution environment uses Python~3.12, NumPy~1.26.4, and pandas~2.2.3, with a dependency configuration that differs from the original evaluation container.

\paragraph{Problem-statement correction.}
\label{app:statement-correction}

The original problem statement for \texttt{django\_\_django-13513} was inconsistent with the actual test objective and has been corrected.
Both the baseline reevaluation and the incremental run for this instance use the same corrected statement, and both pass evaluation; the correction and input-verification records are retained in the SWE-bench Verified study archive.

\subsection{Reported Evidence and Scope of Inference}

This appendix reports the archived cohort outcomes, the diagnoses for $38$ originally failing runs, and the execution results and boundaries of the selected local replays.
The analysis draws on the study archive's trajectories, patches, evaluation logs, and replay results; Appendix~\ref{app:artifact} lists the contents of the submitted review package.
Sphinx builds use an isolated Python~3.12.3 environment with pinned dependencies.

These instances confirm specific evaluation blind spots in another benchmark and coding framework.
The two datasets use the same model, while tasks, run budgets, and frameworks change together; the independent causal effect of architecture requires controlled experiments.
Within the $500$-run collection, our analysis covers human-reviewed evidence-based diagnoses of the $38$ originally failing runs, including two terminal propagation omissions (one with a local repair replay), and execution-based validation of three purposively selected successful patches.
The scope of inference is defined in Section~\ref{sec:discussion}.
We executed both the official evaluation of the original model patches and the program replays before and after consumer synchronization, retaining inputs, patches, environments, and outputs separately.
The original evaluation columns in Table~\ref{tab:sweverified-replays} correspond to the original model patches, whereas the local synchronization results correspond to the target checks and controls listed for each case.
The original $62$ evaluation tests for Sphinx~9461 were also executed separately before and after consumer synchronization.

\section{Terminal Mismatches and Recovery on FeatureBench with Qwen3.8-Flash}
\label{app:featurebench}

\subsection{Selection, Run Settings, and Evidence Method}

This extension uses archived FeatureBench tasks \citep{featurebench2026} run with
\texttt{openai/qwen3.8-flash} and \texttt{mini-swe-agent 2.4.6}; the prefix is the framework's provider tag.
Of the original $100$ tasks, $46$ completed an answer with status \texttt{Submitted}, and the remaining $54$ exceeded the answer-generation time limit with status \texttt{TimeExceeded}.
The adjudicated case analysis in this appendix is limited to these $46$ completed first attempts across $15$ repositories, fixed before retries.
\texttt{Submitted} records completion of the agent's answer, not passage of the task evaluation:
the original evaluator marks $9$ of these $46$ runs resolved and $37$ failed.
The supplementary full-archive failure-mode analysis, including the other $54$ runs, appears in Appendix~\ref{app:featurebench-modes}.
This subset does not support a population mismatch-rate estimate or a model comparison.

We manually reviewed the task requirements, source shown in the archived trajectories, message-level tool calls and outputs, and final patches, and adjudicated the three terminal diagnoses and two additional recovery cases.
The DeepSWE three-reviewer blinded procedure and its $97\%$/$89\%$ agreement figures do not apply to this extension.
We manually confirmed terminal diagnoses in three runs: Meson dependency naming, MLflow host patterns, and Seaborn Path styling.
The first two consumer relations now have supplementary execution support; the third has execution support from the archived original evaluation.
Four separate runs contain five execution-supported recovery events: Seaborn plotting, xarray tree nodes (two events), xarray chunk alignment, and scikit-learn nested validation.
Human adjudication confirmed the latter two recovery cases while retaining the scope and classification boundaries described below.
The recovery witnesses and Seaborn terminal witness come from the original trajectories or archived evaluations; the revised Meson and MLflow terminal labels incorporate the authors' confirmation of completed supplementary executions. These additions do not establish repair-replay outcomes.
The shared framework version and different model/benchmark support additional instances, without isolating model or framework effects.

All message references below are zero-based indices in \path{traj.json}; patch line numbers are physical lines in \path{final_patch.diff}.

\subsection{Three Human-Confirmed Terminal Diagnoses}

\begin{table}[ht]
\centering\small
\caption{Three FeatureBench terminal diagnoses. Evidence labels refer to the diagnosed existing consumer path. Original evaluation failures do not by themselves establish execution evidence for that path. All three runs fail the original evaluation.}
\label{tab:featurebench-terminal}
\begin{tabularx}{\linewidth}{p{.14\linewidth}p{.18\linewidth}Yp{.14\linewidth}}
\toprule
Task & Final F2P; P2P & Retained consumer relation & Evidence \\
\midrule
Meson\newline Cargo & $13/37$; $103/103$ & New dependency names and wrap registrations disagree with existing resolver queries & Execution \\
MLflow\newline security & $28/30$; $94/94$ & New default IPv6 patterns misfeed the shared matcher used by existing FastAPI host validation & Execution \\
Seaborn\newline Path styles & $28/29$; $199/199$ & Shared capstyle adaptation discards the dashed-path theme setting & Execution \\
\bottomrule
\end{tabularx}
\end{table}

\paragraph{Meson: dependency registration and resolver lookup.}
The task is \path{mesonbuild__meson.f5d81d07.cargotests.8e49c2d0.lv1}.
$\Delta$ is the new \texttt{\_dependency\_name} and Cargo-lock wrap generation;
$O$ is agreement between registered dependency names and the names queried by existing resolution paths;
$C$ is \texttt{Interpreter.\_fetch\_package} $\to$ \texttt{\_fetch\_package\_from\_subproject} $\to$ \texttt{find\_dep\_provider}.
$D$ is the mismatch between a normalized name such as \texttt{foo\_1-rs} and the wrap's registered \texttt{foo-1-rs}, now supported by supplementary execution.
The git branch likewise registers package/API names while the existing \texttt{\_dep\_package} path queries repository-directory names.
Messages $7$, $39$--$40$, $62$, $107$, and $139$ locate normalization, callers, registration, and lookup; messages $141$ and $143$ display parser/wrap outputs.
Final-patch lines $109$--$128$ and $171$--$173$ retain these naming decisions.
We treat the related naming manifestations as one diagnosed run, without counting them as independent samples.
The original Cargo test failures are separate from this consumer diagnosis. Supplementary checks now provide execution support for the resolver consumer relation; the original evaluation counts above are retained.

\paragraph{MLflow: default IPv6 patterns and existing host validation.}
The task is \path{mlflow__mlflow.93dab383.test_security.98fcde1d.lv1}.
$\Delta$ is generation of a default \texttt{[::1]:*} pattern together with shell-style matching;
$O$ is to preserve acceptance of the intended localhost-with-port values when defaults reach existing host validation;
$C$ is \texttt{HostValidationMiddleware.\_\_call\_\_} in \path{mlflow/server/fastapi_security.py};
$D$ is the mismatch between a literal IPv6 bracket and the bracket syntax consumed by \texttt{fnmatch}.
Messages $7$ and $9$ retain the original localhost constants and FastAPI consumer.
Final-patch lines $203$--$229$ generate defaults and $232$--$260$ define matching.
The archived evaluation specifically expects the escaped default \texttt{[[]::1]:*} and observes the helper returning false for \texttt{[::1]:8080}.
The original executions confirm the default/helper defect; supplementary checks now provide execution support for the existing middleware consumer relation. The original evaluation counts above are retained.

\paragraph{Seaborn: shared adaptation loses an existing Path theme setting.}
The task is \path{mwaskom__seaborn.7001ebe7.test_line.25304a8b.lv1}.
$\Delta$ is shared \texttt{Mark.\_handle\_capstyle} adaptation;
$O$ is preservation of the separate dashed-line capstyle supplied by a plot theme;
$C$ is the existing \texttt{Path.\_plot} path constructing a Matplotlib \texttt{Line2D};
$D$ is an executed dashed plot whose capstyle is \texttt{butt} instead of the requested \texttt{round}.
Message $6$ shows the original consumer; final-patch lines $138$--$152$ retain the adapter.
It falls back to \texttt{lines.solid\_capstyle='projecting'} and maps the dashed/projecting combination to \texttt{dash\_capstyle='butt'}, losing \texttt{lines.dash\_capstyle='round'}.
The original study-archive file \path{tests/_marks/test_line.py}, lines $124$--$132$, executes
\texttt{Plot(...).add(Path(linestyle='--')).theme(rc).plot()};
\path{evaluation_logs/test_output.txt}, lines $14$--$20$, records the resulting \texttt{Line2D} assertion failure.
This execution witness concerns theme settings, not the later explicit artist-keyword override check that the failed test never reaches.

\subsection{Four Runs with Recovery in the Original Trajectories}

\begin{table}[ht]
\centering\small
\caption{Four archived FeatureBench recovery runs (five events). F2P/P2P are passing counts from the original final evaluation; all four runs remain unresolved. Recovery checks were executed by the original agent and are separate from the final evaluation.}
\label{tab:featurebench-recovery}
\begin{tabularx}{\linewidth}{p{.14\linewidth}p{.18\linewidth}YY}
\toprule
Task & Final F2P; P2P & Consumer failure in the trajectory & Archived recovery evidence \\
\midrule
Seaborn\newline plotting & $196/202$; $245/245$ & Coordinate removal makes \texttt{Bar.\_make\_patches} raise \texttt{KeyError: 'x'} & Same plotting script succeeds after two guards; $536$ passed, $3$ skipped, $1$ xfailed \\
xarray\newline tree nodes & $33/34$; $40/40$ & Detachment invalidates a live dictionary view, then rejects the existing delete-before-orphan path & Dictionary rebinding and removal of the strict key-count check; $153$ passed, $5$ xfailed \\
xarray\newline chunks & $14/18$; $101/101$ & Early return leaves overlapping Zarr writes; readback loses the last row & Remove single-backend-chunk early return; $6$ passed, $3$ skipped \\
scikit-learn\newline validation & $215/218$; $788/788$ & New wrapper's validation context rejects the existing internal LARS call & Skip nested validation as on the estimator path; $37$ passed \\
\bottomrule
\end{tabularx}
\end{table}

\subsection{Seaborn: Coordinate Handling Breaks an Existing Bar Consumer}

The task identifier is \path{mwaskom__seaborn.7001ebe7.test_plot.b645d353.lv1}.
Its statement requests restoration of declarative plotting interfaces, including paired variables and transformations.
For $e=(\Delta,O,C,D)$, $\Delta$ is the edit to existing coordinate handling in
\texttt{Plotter.\_compute\_stats} and \texttt{\_plot\_layer}: an existing coordinate column is dropped before renaming a paired variable.
$O$ is to preserve coordinates already produced by a statistical transform when its output no longer contains the original paired-variable column.
$C$ is the existing \texttt{Bar.\_plot} $\to$ \texttt{Bar.\_make\_patches} path, which reads \texttt{data[other]} to construct bars.
$D$ is the same plotting script's success, failure at this consumer after the edit, and success after guarded synchronization.

\paragraph{Initial state and failure.}
Message $137$ creates synthetic data using \texttt{RandomState(0)}, generating \texttt{penguins} before \texttt{tips}; reproducing the inputs requires that generation order.
Messages $145$ and $149$ construct and execute \path{/tmp/work/t2.py}, including
\texttt{Plot(tips, y="total\_bill").pair(x=["size", "day"]).add(Bar(), Hist())} followed by \texttt{plot()}.
Message $150$ reports \texttt{stat+pair ok 2}.
This is an earlier working state within the trajectory, not an execution of the untouched FeatureBench starting repository.
Message $155$ applies the coordinate edit; message $156$ reports \texttt{KeyError: 'x'}.
Messages $159$--$160$ rerun the same script and retain the traceback through \texttt{\_plot\_layer}, \texttt{Bar.\_plot}, and \texttt{Bar.\_make\_patches};
the latter reads a coordinate column already removed upstream.

\paragraph{Recovery and final-patch binding.}
Message $163$ guards both coordinate-removal sites by requiring the paired-variable column to exist as well:
\texttt{var in df and axis in df}, with the corresponding condition on \texttt{out\_df}.
Message $164$ records \texttt{stat+pair ok 2} from the same script and $536$ passed, $3$ skipped, and $1$ xfailed in the agent's core/mark/stat tests.
Final-patch lines $1059$--$1078$ retain both guards (conditions at lines $1066$ and $1076$).
This is one recovered consumer relation, not a terminal omission.
The original final evaluation still has six F2P failures (Table~\ref{tab:featurebench-recovery}); recovery of this relation does not imply task success.

\subsection{xarray: Restored Detachment Conflicts with Two Existing Deletion Paths}

The task identifier is \path{pydata__xarray.97f3a746.test_treenode.aa8ba777.lv1}.
Its statement requests restoration of tree-node operations, including parent reassignment and orphaning.
Messages $6$ and $8$ preserve the original consumers before the new detach implementation:
\texttt{TreeNode.children.deleter} iterates over \texttt{self.children.values()} and calls \texttt{child.orphan()}, whereas
\texttt{TreeNode.\_\_delitem\_\_} deletes a mapping key before calling \texttt{child.orphan()}.
The \texttt{DataTree} delegation is also recorded in the trajectory.
Message $146$ creates the first implementation block for \texttt{\_detach\_child}; message $158$ installs it in \path{xarray/core/treenode.py}.
It deletes dictionary entries in place and requires exactly one key for the child.
Here the consumers predate the edit, but there is no pre-edit successful execution witness for the restored operation.

\paragraph{Event 1: preserving an active consumer's iteration.}
$\Delta$ is in-place dictionary deletion by the new detach operation;
$O$ is to preserve the old mapping view while the existing bulk-deletion consumer iterates through it;
$C$ is \texttt{children.deleter}, reached through the setter and \texttt{orphan()};
$D$ is the runtime failure of the existing \texttt{test\_is\_hollow} test.
The original source at message $6$ shows this consumer at lines $186$--$194$.
Messages $186$--$187$ execute the DataTree tests and record
\texttt{RuntimeError: dictionary changed size during iteration} at this path ($1$ failed, $17$ passed).
Message $188$ changes detachment to rebind \texttt{self.\_children} to a filtered new dictionary.
Message $189$ advances beyond that test to the second consumer failure; message $191$ subsequently confirms the whole test file passes apart from expected failures.

\paragraph{Event 2: respecting delete-before-orphan ordering.}
$\Delta$ is the same operation's strict one-key precondition;
$O$ is to accept the already-deleted-key state produced by an existing caller;
$C$ is \texttt{TreeNode.\_\_delitem\_\_}, shown at original source lines $649$--$656$ in message $8$;
$D$ is the existing \texttt{test\_delitem} failure in message $189$, reporting zero keys where one was required ($1$ failed, $55$ passed, $4$ xfailed).
Message $190$ removes the strict key-count check while retaining dictionary rebinding.
Message $191$ reports $153$ passed and $5$ xfailed for \path{xarray/tests/test_datatree.py}.
Final-patch lines $232$--$238$ retain both changes.
The two obligations belong to distinct existing consumer paths within one file and count as two recovered events in one run, not two independent samples.
The final evaluation still fails the duplicate-child F2P test: an \texttt{AssertionError} occurs where \texttt{InvalidTreeError} is required ($33/34$ F2P, $40/40$ P2P), separate from these recovered relations.

\subsection{Two Additional Human-Confirmed Recovery Cases}

\paragraph{xarray: chunk alignment and an existing Zarr writer.}
The task is \path{pydata__xarray.97f3a746.test_backends_chunks.fa55f68a.lv1}.
$\Delta$ is the early return in the restored \texttt{\_align\_nd\_chunks}, introduced in message $44$, that leaves variable chunks unchanged for a single backend chunk;
$O$ is to preserve stored values by keeping concurrent writes to a storage chunk from overlapping;
$C$ is the existing \texttt{Dataset.to\_zarr} and Zarr backend path with \texttt{align\_chunks=True};
$D$ is executed data corruption followed by recovery after removal of that early return.
Message $101$ records six failed Zarr alignment tests and three skips.
For a $2\times2$ array, message $109$ observes readback \texttt{[[0,1],[0,0]]} instead of \texttt{[[0,1],[2,3]]}; message $113$ provides a successful single-Dask-chunk control.
Message $114$ removes the single-backend-chunk early return, and message $117$ reports the same group with six passed and three skipped.
Final-patch lines $300$--$303$ retain only the single-variable-chunk early return, preserving the recovery.
We manually confirmed this recovered consumer relation while retaining a task-scope conflict: message $1$ explicitly requires dimensions with one backend chunk to be returned unchanged.
The recovery addresses the existing writer's data-integrity obligation but changes that stated special case; confirmation does not make the scope undisputed.
The final evaluation retains four direct chunk-grid failures ($14/18$ F2P, $101/101$ P2P).

\paragraph{scikit-learn: a restored wrapper changes nested validation.}
The task is \path{scikit-learn__scikit-learn.5741bac9.test_public_functions.28421aef.lv1}.
$\Delta$ is the restored \texttt{graphical\_lasso} wrapper with \texttt{prefer\_skip\_nested\_validation=False} in messages $181$ and $183$;
$O$ is to preserve the validation context required by the existing internal LARS call when exposing the algorithm through the public wrapper;
$C$ is \texttt{\_graphical\_lasso} $\to$ \texttt{lars\_path\_gram(method="lars")}, shown before the wrapper edit in message $103$;
$D$ is an executed \texttt{InvalidParameterError} and recovery under the estimator's validation context.
Message $186$ records five failed and $32$ passed covariance tests; message $188$ shows nested validation rejecting \texttt{method='lars'}.
Message $190$ runs the successful \texttt{GraphicalLasso(mode='lars').fit} control.
Message $191$ changes only the wrapper's nested-validation setting to \texttt{True}; message $192$ reports $37$ passed.
Final-patch lines $1279$--$1288$ retain that setting, and lines $1524$--$1533$ retain the direct internal call.
We manually confirmed the recovery as an execution-context dependency on an existing consumer.
The classification boundary is that the repair lies in the newly restored wrapper's own configuration; this is distinct from the explicit task-scope conflict in the xarray case.
The original final evaluation still has three wrapper/estimator parameter-validation failures ($215/218$ F2P, $788/788$ P2P), separate from the recovered LARS path.

\subsection{Evidence Scope}

The case descriptions above provide the task identifiers, edit--obligation--consumer--diagnosis chains, message and patch coordinates, original evaluation outcomes, and the distinction between archived and supplementary execution support for the three terminal diagnoses and four recovery runs.
The coordinates identify the original trajectories and final patches in the FeatureBench study archive; evaluator logs and test sources explain the evidence for each case.
Appendix~\ref{app:artifact} lists the contents of the submitted review package.

\section{Failure Modes and Diagnostic Boundaries in the Full FeatureBench Archive}
\label{app:featurebench-modes}

\subsection{Coverage, Evidence Linkage, and Counting Units}

Appendix~\ref{app:featurebench} details the human-adjudicated relations; this section adds \emph{failure modes visible in the archived materials} for all $100$ original first attempts.
The analysis combines original evaluation logs, test sources, task-cropping/restoration patches, and model trajectories and final patches linked by source manifests from the two supplementary evidence packages, \texttt{submitted46} and \texttt{timeout54}.
The two sets do not overlap and together cover $18$ repositories. The SHA-256 hashes of all $100$ final patches and $100$ trajectories match the supplementary packages' source records, and all $14{,}938$ files in the two package manifests pass hash verification.
The exported per-task evaluation objects agree item by item with the scores and test statuses in the source records; hashes of original report files and file hashes of exported objects are handled separately.
This section extends archive analysis for the additional candidate relations. Existing human adjudications retain their original scope; no additional executions or consumer-repair replays were performed for these candidates. Supplementary execution support for the adjudicated terminal diagnoses is reported separately in Appendix~\ref{app:featurebench}.

\begin{table}[htbp]
\centering\small
\caption{Answer status and evaluation records for the original FeatureBench runs. The F2P/P2P columns indicate whether the original report's failure lists are nonempty, rather than directly indicating confirmed code defects. The five outcome categories in each row are mutually exclusive.}
\label{tab:featurebench-full100}
\begin{tabularx}{\linewidth}{Yrrrrrr}
\toprule
Answer status & Passed & \shortstack{F2P-only\\failures} & \shortstack{P2P-only\\failures} & Both failed & \shortstack{No results\\listed} & Total \\
\midrule
\texttt{Submitted} & $9$ & $34$ & $2$ & $1$ & $0$ & $46$ \\
\texttt{TimeExceeded} & $13$ & $33$ & $0$ & $3$ & $5$ & $54$ \\
\midrule
Total & $22$ & $67$ & $2$ & $4$ & $5$ & $100$ \\
\bottomrule
\end{tabularx}
\end{table}

\texttt{TimeExceeded} means that the model's answer generation reached its time budget; it does not imply that the saved patch necessarily failed evaluation, nor does it mean that an individual test timed out.
Of these $54$ records, $13$ still pass the original evaluation; examining only normally submitted answers would omit this group.
Another $5$ records fail the original scoring but have empty F2P and P2P results: $3$ have empty patches, and the other $2$ are blocked during import. Empty failure lists therefore cannot be read as all tests passing.
The table retains the original evaluator's total of $78$ nonpassing runs, without rescoring them according to this section's diagnoses.

For each run, we separately record the original score, observed failures, relevant patch/cropping sources, relation classification, and evidence coordinates.
A run may contain implementation omissions, local algorithm errors, environmental interference, and recovered events at the same time; the modes below are therefore nonexclusive.
Multiple parametrized failures triggered by the same missing interface do not count as independent propagation relations, and recovery events do not count as omissions in the final patch.
The $22$ runs that pass the original scoring are also included in the index. Their status means only that the original evaluator marked them as passing, and does not establish the absence of untested defects.

\subsection{Mismatch Mechanisms Along Consumer Relations}

Table~\ref{tab:featurebench-propagation-modes} groups the previously adjudicated $3$ terminal-diagnosis runs and $5$ recovery events in $4$ runs by consumer obligation.
These mechanisms involve keys, representations, styles, data columns, shared containers, operation ordering, storage layouts, and implicit validation contexts; function call graphs or final test scores alone cannot distinguish them.

\begin{table}[htbp]
\centering\small
\caption{Eight concrete mechanisms in adjudicated FeatureBench events. All three terminal diagnoses have execution support, including supplementary support for Meson and MLflow; recovery witnesses follow Appendix~\ref{app:featurebench}. Mechanism counts differ from run counts.}
\label{tab:featurebench-propagation-modes}
\begin{tabularx}{\linewidth}{p{.23\linewidth}Yp{.19\linewidth}}
\toprule
Mismatch mechanism & Unmet existing consumer obligation & Record and state \\
\midrule
Unsynchronized registration and lookup keys & Hyphen normalization of package/API names, wrap naming, and provider lookup must use a consistent identity & Meson; terminal/execution \\
Confusion between literal representation and pattern syntax & IPv6 brackets must be escaped before reaching the shared wildcard matcher & MLflow; terminal/execution \\
Shared adapter overrides consumer defaults & Path's dashed-line capstyle must preserve the theme's round setting & Seaborn Line; terminal/execution \\
Premature deletion of intermediate data columns & Coordinates produced by paired statistics must remain available to the existing Bar consumer & Seaborn Plot; recovered \\
In-place mutation breaks iteration safety & Child detachment must not change the size of a dictionary view being iterated & xarray tree nodes; recovered \\
Precondition conflicts with existing call order & The delete-before-detach path must allow the key to be already absent & xarray tree nodes; recovered \\
Compute chunks conflict with storage-write obligations & Multiple uncoordinated tasks must not separately overwrite one Zarr storage chunk & xarray chunks; recovered/disputed scope \\
Wrapper changes implicit call context & The public entry point must preserve the nested-validation setting needed by the internal LARS call & sklearn validation; recovered \\
\bottomrule
\end{tabularx}
\end{table}

\paragraph{Additional candidate relations in the timed-out group.}
Joint inspection of the timed-out group's patches and original execution stacks identifies the following consumer relations.
We list $6$ additional types of archived candidate relations across $5$ runs, giving the edit, obligation, existing consumer, and failure witness for each. These are neither added to the previous $3$ human-confirmed terminal diagnoses nor used to estimate a full-corpus propagation mismatch rate.

\paragraph{Broadened semantics of lazy-loading state.}
In Astropy's \texttt{test\_tilde\_path}, the newly implemented \texttt{\_BaseHDU.\_has\_data} changes ``loaded and nonempty'' to ``loaded or positive on-disk data length'' (final-patch lines $863$--$867$; cropping-patch lines $1734$--$1739$ preserve the original contract).
The existing table HDU's \texttt{columns} property interprets a true value as permission to read cached \texttt{data}.
Execution of \texttt{fits.printdiff} produces recursion along \texttt{data}$\to$\texttt{\_get\_tbdata}$\to$\texttt{columns}$\to$\texttt{data}; original-log lines $77$--$113$ record a \texttt{RecursionError}.
The obligation is to preserve the loading-stage meaning of the state predicate for its consumer. Other failures in the same task due to the missing \texttt{\_get\_file\_mode} are classified separately; not all $33$ failures in this task can be attributed to recursion.

\paragraph{Parsing loses the missing-value mask.}
Within the \texttt{BitArray} class in Astropy's \texttt{test\_vo}, task cropping removes only \texttt{\_splitter\_lax}, but the model additionally introduces \texttt{BitArray.parse}, overriding the inherited parsing logic (final-patch lines $15$--$39$).
It pads short values with False and returns \texttt{mask=False} throughout that branch.
The existing \texttt{BitArray.output/binoutput} relies on the mask to distinguish missing bits from actual zeros and issue a W39 warning accordingly.
Original-log lines $312$--$330$ record a mask assertion failure, and round-trip serialization also writes empty cells as strings of zeros. The missing expected W39 warning in the Binary2 fixture leads to further setup errors.
The candidate obligation is to preserve the missing-value metadata needed by the consumer during parsing; multiple errors caused by the shared fixture do not count as additional propagation relations.

\paragraph{Omitted normalization prerequisite.}
In Seaborn's \texttt{test\_scales}, the new \texttt{ContinuousBase.\_setup} performs normalization only when \texttt{norm} is explicitly set (final-patch lines $560$--$577$), then still passes the result to the existing interval-mapping closure in \texttt{IntervalProperty.get\_mapping}.
Neither task cropping nor the final patch modifies this closure, whose linear mapping assumes normalized inputs.
The default input $[1,3,9]$ should first become $[0,0.25,1]$, but the original execution still yields $[1,3,9]$. With a target interval specified, it yields $[3,7,19]$ instead of $[1,1.5,3]$ (original-log lines $35$--$64$).
This provides a candidate relation in which the values remain computable, but the producer and existing consumer use different scales.
Because normalization is also a direct contract of the new \texttt{\_setup} itself, this example retains the classification boundary between a local algorithm error and a cross-stage mismatch.
Other errors in the same task involve RGB/RGBA shapes, transform-name parsing, and locator/formatter initialization, and should be recorded separately.

\paragraph{Empty-value normalization and backend dispatch.}
The new \texttt{parse\_email} in Packaging's \texttt{test\_metadata} stores an empty \texttt{Import-Name} as \texttt{['']}. The existing \texttt{\_Validator.\_process\_import\_names} then treats the empty string as an identifier and rejects it; original-log lines $18$--$40$ show that an empty list is expected.
The candidate obligation is to convert the representation of ``no name supplied'' into an empty collection accepted by the consumer.
In Pandas' \texttt{test\_reduction}, the backend decorators attached to the new \texttt{nanany/nanall} and the \texttt{\_bn\_ok\_dtype} decision lead into the existing \texttt{bottleneck\_switch}, which attempts to call a nonexistent backend function and raises an error because a \texttt{NoneType} value is not callable.
The candidate obligation is to satisfy both dtype compatibility and backend-operation availability. Because the new functions and decorator attachment are also part of the patch, a local configuration error must still be distinguished from an independent consumer-relation mismatch.

\paragraph{Bypassing the compressed-stream wrapper.}
Astropy's \texttt{test\_vo} contains another interface candidate: the new XML input adapter uses the availability of \texttt{fileno} to decide whether to pass a file object directly, whereas the original contract takes this branch only for raw \texttt{io.FileIO} and supplies a \texttt{read} callback for other objects (final-patch lines $1064$--$1076$; cropping-patch lines $1284$--$1291$).
The original compressed-file test reports \texttt{invalid token}, consistent with the static chain in which the downstream parser bypasses the decompression wrapper.
The log does not directly observe the bytes received downstream, so this remains an interface-adaptation candidate rather than a causal mechanism confirmed by replay.

\subsection{Ordinary Implementation Failures and Gaps Left by Task Cropping}

Table~\ref{tab:featurebench-other-modes} lists other modes in the full archive that need to be distinguished from propagation diagnoses.
Classification uses actual exceptions, assertion differences, and patch contents rather than test names alone.
For example, ``the expected ValueError message differs'' and ``invalid input is not rejected'' occur at different levels; an array that fails to meet requirements also does not automatically establish a mismatch between two independent source locations.

\begin{table}[htbp]
\centering\small
\caption{Other failure modes in the full archive. Each row summarizes several concrete subtypes; one run may exhibit multiple rows' modes, which are not treated as mutually exclusive primary-cause frequencies.}
\label{tab:featurebench-other-modes}
\begin{tabularx}{\linewidth}{p{.23\linewidth}YY}
\toprule
Mode and subtypes & Representative archived observations & Boundary with propagation diagnosis \\
\midrule
Unrestored implementation/unconnected symbols & Missing MLflow \texttt{\_get\_store} and \texttt{call\_endpoint}, Seaborn \texttt{assign\_variables}, and Sphinx \texttt{find\_uri} & Failure to restore definitions or imports already removed by input cropping does not mean that the model's edit creates a new consumer obligation \\
Input-validity and type-selection errors & FastAPI treats fields that should be rejected as scalar; duplicate TreeNode children trigger an internal assertion; sklearn parameter validation fails & Decisions, boundaries, and error exits in the new logic do not automatically count as propagation \\
Incompatible diagnostic contracts & Different exception messages/types, warning messages/categories, log levels, or command-line argument-name ordering; examples in RGB, MLflow, and Sphinx & The observable interface may be violated even when the main algorithm runs; specify whether the failure concerns type or text \\
Representation, formatting, and default-precedence errors & Meson version ranges/URLs; bytes field names incorrectly converted to strings; Pydantic-generated titles override explicit titles; missing HTML colons and serialization case/newline errors & First locate the new implementation's output contract; discuss propagation only if a retained consumer has an additional independent obligation \\
Numerical, shape, and layout errors & Histogram zero bins/lost tail samples; Bars array-shape mismatches; RGB/RGBA channel counts; rechunk layouts and dtype/index names & Use actual numerical, shape, and metadata assertions rather than calling every error an algorithmic capability deficit \\
State, initialization, and resource-protocol errors & Unbound locators/formatters; subprocess-wrapper protocols, path handling, and resource-input adaptation errors & Distinguish local initialization, external-library interfaces, and existing consumers within the repository \\
Mismatch in randomness and observable protocols & Telemetry tests control \texttt{random.randint}, but the patch uses \texttt{random.random}; thread names differ & Original assertions do not establish incorrect sampling mathematics or concurrent causality; retain the tested control-interface difference \\
Environmental and run-condition interference & Expired Astropy leap-second tables cause time-conversion warnings; skips for missing optional dependencies are recorded separately & P2P failures may also be environmental; without before/after controls in the original environment, edit-induced causation is not established \\
No evaluable artifact/import blocking & $3$ empty patches; Pandas concat and xarray coordinate-transform imports are interrupted & Missing test results differ from all tests passing; specific behavior cannot be inferred from absent results \\
Confused test-log levels & Expected failures nested inside pytest self-tests are recorded as outer P2P failures & This is an evaluation-parsing issue; retain the original score and analyze program attribution separately \\
\bottomrule
\end{tabularx}
\end{table}

\paragraph{Parsing, boundaries, and uniqueness.}
The per-task index retains distinct local mechanisms within each broad category: sklearn ARFF removes quotes twice at two layers, losing literal quotes inside strings; Sphinx C macro parsing incorrectly accepts variadic parameters in the middle of a parameter list; node-ID generation detects a name collision but does not recheck the fallback name for uniqueness; linkcheck with a zero retry limit still returns a timestamp in its first iteration; and command-line defaults store an empty string in a field that should be \texttt{None}.
These issues respectively concern the number of representation conversions, syntactic-position constraints, global uniqueness, stopping boundaries, and empty-value semantics. A shared exception name cannot replace mechanism analysis.

\paragraph{Unrestored dependencies are not new propagation.}
The $9$ failures in MLflow's \texttt{test\_evaluation\_dataset} arise from missing \texttt{\_get\_store} and \texttt{TraceData.\_get\_root\_span}; the $12$ failures in \texttt{test\_rest\_store\_webhooks} involve missing \texttt{message\_to\_json}/\texttt{call\_endpoint}.
The input cropping patch already deletes these definitions or imports, and the model's final edits modify other interfaces without restoring them.
Similarly, the $257$ failures in Seaborn's \texttt{test\_distributions} are attributable to missing \texttt{remove\_na} and \texttt{assign\_variables}, while the $105$ failures in \texttt{test\_relational} concentrate on four unrestored members/helpers.
These results support ``failure to restore dependencies left by task cropping''; each failed call cannot be recorded as a new mismatch created by the agent's edits.
For records in which the task statement and cropping coverage differ, the per-task index also retains task-scope notes. This section does not infer that the benchmark's overall construction is invalid.

\paragraph{Failed-test counts are not mechanism counts.}
The $6$ failures in Astropy RGB arise from two exception messages; the $4$ parametrized failures for Pydantic titles correspond to one explicit-title precedence issue.
A shared fixture, missing import, or incorrect representation can multiply into several failures.
Conversely, one xarray tree-node run contains recovery events on two distinct existing deletion paths.
This section therefore records runs, failed test nodes, and consumer relations separately, rather than using failed-test counts as propagation-event counts.

\subsection{Scoring-Parser Issues, Review Entry Points, and Inference Scope}

\paragraph{A P2P failure in the original report need not be a regression.}
Lines $15$--$37$ of \texttt{test\_output\_p2p\_approx.txt} for \texttt{pytest:test\_local} come from a subprocess captured by a successful outer self-test: the inner test deliberately asserts \texttt{[3] == [pytest.approx(4)]} and fails; the outer test then checks its output, and the log ends with $90$ tests passing.
The original evaluator nevertheless lists the nested \texttt{test\_foo} as a P2P failure.
Archived \path{harness/test_parsers.py}, lines $29$--$47$, selects the first \texttt{short test summary info} and stops at the next separator; \path{harness/report.py} then constructs scoring sets from the parsed keys.
Applying this archived parser to the original log yields only the inner \texttt{test\_foo: FAILED} mapping, without the outer $90$ results.
This is a parsing check of existing text, not a rerun of the original task.
The P2P record for \texttt{pytest:test\_raises\_group} in the timed-out group has the same kind of nested-result confusion.
Both tasks also have original F2P failures; this section therefore does not change their nonpassing scores.

\paragraph{Insufficient evidence must also be retained.}
For example, the sole P2P error in Astropy's \texttt{test\_quantity\_erfa\_ufuncs} is \texttt{Time - float} during IERS time conversion, and the same original trajectory also contains an expired-leap-second exception.
The available materials do not uniquely attribute this observation to the model's unit-handling edits, so the unresolved localization and environmental confounding are retained.
``Coverage of all archived runs'' means that each run has an evidence index; it does not mean that every root cause is causally confirmed or that existing tests can detect every potential defect.

\paragraph{Study-archive records.}
The separate FeatureBench study archive contains the per-task audit records, their aggregate statistics, and the parsing-check documentation.
Each record uses the full \texttt{instance\_id} to link answer status, original F2P/P2P nodes, final patch, input cropping, and log line numbers. Appendix~\ref{app:artifact} lists the files in the submitted data supplement.
This section expands archive-analysis coverage while preserving the distinction between previous human confirmations and the present candidates.
All recoveries still occur within the original trajectories, all original scores retain their original values, and no execution claim is made about the benefits of local synchronization for the additional relations.

\section{Additional Related Work}
\label{app:additional-related-work}

\subsection{Comparison with Refactoring and Feature-Development Benchmarks}
\label{app:benchmark-comparison}

The diagnosed program defects overlap with established incomplete change propagation, incomplete migration, and regression failures. The following comparisons clarify how our audit connects specific consumer relations to trajectory evidence and evaluation outcomes.

\paragraph{RefactorBench: multi-file refactoring and state tracking.}
RefactorBench analyzes omitted edit locations even when descriptive instructions identify the relevant files, strict linting that blocks necessary intermediate edit states, and feedback loops that displace the original objective (Section~5). Its state-aware interface presents cumulative edits and selected earlier observations to support state tracking (Section~6.1 and Appendix~G)~\citep{gautam2025refactorbench}.
These findings already connect multi-file coordination to an agent's understanding of the evolving state. Our focal analysis links the presentation of a particular consumer's code to explicit judgments about its adaptation obligation, subsequent judgments about the same obligation, and the resulting synchronization state (Section~\ref{sec:cases}; Appendix~\ref{app:prior-judgment}). This supplies relation-specific evidence about how an obligation is assessed during continued review.

\paragraph{FeatureBench: feature development and interface grounding.}
FeatureBench evaluates complex feature development using new acceptance tests and existing regression tests. Its failure analysis identifies missing cross-file references that produce unresolved symbols or \texttt{NameError}, and interface or attribute guessing that produces \texttt{TypeError} or \texttt{AttributeError} (Section~4.1.3). It also studies interface information, visible tests, and execution-step budgets (Sections~4.2.1--4.2.2)~\citep{featurebench2026}.
Our audit identifies the edit-induced semantic obligation of a specific existing consumer and records its intermediate recovery and terminal state alongside the original evaluation outcome. This includes type-valid semantic loss, such as Kysely's omitted node field, for which fixed-patch consumer checks and local synchronization expose a gap in an originally passing patch (Section~\ref{sec:replay}).

\paragraph{SWE Refactor Bench: migration completeness and behavioral preservation.}
SWE Refactor Bench separately evaluates whole-repository stack-migration completeness and behavioral preservation, including post-submission agent-based differential testing beyond its fixed test suite (Section~2.3). Its analysis distinguishes incomplete migrations from behavior-breaking migrations and discusses downstream consumers and counterexamples missed by fixed tests (Sections~3.3--3.4)~\citep{hong2026swerefactorbenchcoding}.
Our event-level audit connects an applied edit, its adaptation obligation, an existing consumer, and a diagnosis to recovery within the original trajectory, terminal omissions, and original scores. Selected local synchronization replays then test the behavioral consequences of the diagnosed consumer relations. These measurements complement migration-level completeness and correctness assessments.

\paragraph{Additional process evidence and comparison scope.}
In termenv, consumer-code presentation precedes two explicit judgments accepting an unsynchronized renderer; the incorrect judgment is formed after the edit and retained during later review. The supplementary eicrud sequence instead links a tentative pre-edit decision to a later decision retaining it after a new constraint, with disputed task scope. Appendix~\ref{app:prior-judgment} specifies the source fields and evidentiary limits of these records.
The dependency-review intervention tests the practical value of re-examining consumer obligations under matched starting context, code state, and token-budget caps (Appendix~\ref{app:dependency-review}). The comparisons above concern documented findings and measurement units; evaluating the prompt against state-aware interfaces or migration validators would require a shared experimental protocol.

\subsection{Other Repository-Level Approaches}

AutoCodeRover and CodeR use program structure or task graphs to organize edits~\citep{Zhang_2024,chen2024coderissueresolvingmultiagent}.
Aider provides repository maps~\citep{aider2026repomap}.
SWE-Touch tests adaptation to simulated external edits and conflicting messages~\citep{tan2026swetouchbenchmarkingcodingagents}.
MAGIS uses multi-role quality review, while SWE-Search uses tree-search feedback~\citep{NEURIPS2024_5d1f0213,ICLR2025_a1e6783e}.

\section{Supplementary Codex Archive Audit on FeatureBench}
\label{app:codex-featurebench}

\paragraph{Model and selection.}
This extension audits formal baseline archives produced by Codex CLI using \texttt{gpt-5.6-sol} with reasoning effort \texttt{high} on FeatureBench.
Tasks were fixed before the new Codex outcomes were available, excluding simple tasks on which both Codex and Qwen passed all evaluations; the final cohort contains $72$ tasks, with one selected baseline per task.
The archives use the \texttt{fast} split of \texttt{LiberCoders/FeatureBench} at dataset revision \path{76b4a4566e04f4bcc13c35125d4f301791efa736}.
Per-run settings are retained in \path{run_info.json}; archived \texttt{timeout} values vary, so these runs are not treated as a controlled model comparison with a uniform budget.

\paragraph{Audit and evidence.}
The extension is a manual audit of all selected baselines, inspecting original responses, task requirements, final patches, and evaluation records, with manual cross-checks of selected positive and boundary cases by other reviewers followed by adjudication.
It is reported separately from the DeepSWE full-corpus human blinded audit and external expert review. This archival analysis launches no new model runs or program tests; execution witnesses come from tool outputs and evaluation logs saved in the original baselines.
Diagnoses require an applied edit, an adaptation obligation, an existing consumer, and a specific error. Recovery requires a failure--repair--verification chain within the original baseline; terminal omissions remain bound to its final patch. Candidate classifications are not automatically counted as identified mismatches.

\paragraph{Results and interpretation.}
Table~\ref{tab:codex-featurebench} reports terminal, recovered, overlapping, and union states against the same fixed baseline denominator.
``Ever observed'' is the union of terminal and recovered runs, not the terminal-omission rate, and is not directly comparable with the DeepSWE terminal-diagnosis share.

\begin{table}[htbp]
\centering\small
\caption{Mismatch states in the Codex baseline archives. Every row uses $72$ runs as its denominator. Cells report run counts and shares, except overlap, which reports counts only. Terminal and recovered states may overlap; the final column is their union.}
\label{tab:codex-featurebench}
\begin{tabularx}{\linewidth}{Yrrrr}
\toprule
Scope and evidence & Terminal & Recovered & Both & Ever observed \\
\midrule
Execution-supported, undisputed scope & $2$ ($2.8\%$) & $11$ ($15.3\%$) & $1$ & $12$ ($16.7\%$) \\
Execution-supported, including disputed scope & $3$ ($4.2\%$) & $11$ ($15.3\%$) & $1$ & $13$ ($18.1\%$) \\
Also including source-supported terminal diagnoses & $4$ ($5.6\%$) & $11$ ($15.3\%$) & $1$ & $14$ ($19.4\%$) \\
\bottomrule
\end{tabularx}
\end{table}

The strict terminal runs are \texttt{037} and \texttt{039}, both originally nonpassing; \texttt{039} also has a recovered event involving a different consumer relation.
The broader execution-supported set includes \texttt{058}: the old \texttt{mean} path has an execution witness, but the task explicitly excludes general aggregation, leaving adaptation scope disputed.
In \texttt{012}, a bare module name conflicts with the existing \texttt{module://} resolver protocol at source level without an archived execution witness for the downstream import; it enters only the source-supported extension.
These tasks are not a random sample of the full benchmark, and audit methods, selection, run conditions, and statistical objects differ across corpora. Results describe the selected baselines and broaden model coverage without supporting model rankings or general mismatch-probability estimates.

\paragraph{Materials and recomputation.}
Within \path{codex_72_baseline_model_responses/}, \path{index.csv} indexes the tasks, and \path{records/} retains model events, patches, tasks, settings, and original evaluation logs.
The per-run labels are in \path{diagnoses.csv} and \path{diagnoses.jsonl}. The run sets and aggregates for Table~\ref{tab:codex-featurebench} are in \path{counts.json} and can be recomputed from the per-run labels.

\end{document}